\documentclass[review,authoryear,times]{elsarticle}
\usepackage[utf8]{inputenc}  

\usepackage{xcolor}
\definecolor{darkblue}{rgb}{0, 0, 0.5}

\usepackage{latexsym}
\usepackage{amsmath}
\usepackage[hyphens]{url}
\usepackage{graphicx}
\usepackage{adjustbox}
\usepackage{subcaption}
\usepackage{amssymb}
\usepackage{multirow}
\usepackage{boxedminipage}
\usepackage{multirow}
\usepackage{rotating}
\usepackage{xfrac}

\newcommand\Tstrut{\rule{0pt}{2.3ex}}       
\newcommand\Bstrut{\rule[-0.9ex]{0pt}{0pt}} 

\usepackage{mathtools}
\usepackage{xspace}
\usepackage{afterpage}
\usepackage{pdflscape}
\usepackage{array}
\usepackage{arydshln}
\usepackage{float}
\usepackage{dcolumn}

\usepackage{xcolor}
\definecolor{darkblue}{rgb}{0,0,0.5}
\newcommand{\bleu}{BLEU\xspace}
\newcommand{\rouge}{ROUGE-L\xspace}
\newcommand{\meteor}{METEOR\xspace}
\newcommand{\cider}{CIDEr\xspace}
\newcommand{\nist}{NIST\xspace}

\newcommand\tgen{\textsc{TGen}\xspace}
\newcommand\slug{\textsc{Slug}\xspace}
\newcommand\slugalt{\textsc{Slug-alt}\xspace}
\newcommand\tntnlgi{\textsc{TNT1}\xspace}
\newcommand\tntnlgii{\textsc{TNT2}\xspace}
\newcommand\zhawi{\textsc{ZHAW1}\xspace}
\newcommand\zhawii{\textsc{ZHAW2}\xspace}
\newcommand\adapt{\textsc{Adapt}\xspace}
\newcommand\dangnt{\textsc{DANGNT}\xspace}
\newcommand\forgei{\textsc{FORGe1}\xspace}
\newcommand\forgeiii{\textsc{FORGe3}\xspace}
\newcommand\gong{\textsc{Gong}\xspace}
\newcommand\harv{\textsc{Harv}\xspace}
\newcommand\nle{\textsc{NLE}\xspace}
\newcommand\sheffi{\textsc{Sheff1}\xspace}
\newcommand\sheffii{\textsc{Sheff2}\xspace}
\newcommand\chen{\textsc{Chen}\xspace}
\newcommand\thomsoni{\textsc{TR1}\xspace}
\newcommand\thomsonii{\textsc{TR2}\xspace}
\newcommand\tuda{\textsc{TUDA}\xspace}
\newcommand\zhang{\textsc{Zhang}\xspace}

\definecolor{seqtoseq}{rgb}{0.941, 0.062, 0.207}
\definecolor{datadriven}{rgb}{0.960, 0.509, 0.188}
\definecolor{rules}{rgb}{0.156, 0.784, 0.231}
\definecolor{templates}{rgb}{0, 0.4, 0.619}
\newcommand{\symbseq}{$^\heartsuit$}
\newcommand{\symbdd}{$^\diamondsuit$}
\newcommand{\symbrule}{$^\clubsuit$}
\newcommand{\symbtempl}{$^\spadesuit$}

\newcommand\Ctgen{\textcolor{seqtoseq}{\symbseq\bf \tgen}}
\newcommand\Cslug{\textcolor{seqtoseq}{\symbseq\bf \slug}}
\newcommand\Cslugalt{\textcolor{seqtoseq}{\symbseq\bf \slugalt}}
\newcommand\Ctntnlgi{\textcolor{seqtoseq}{\symbseq\bf \tntnlgi}}
\newcommand\Ctntnlgii{\textcolor{seqtoseq}{\symbseq\bf \tntnlgii}}
\newcommand\Czhawi{\textcolor{datadriven}{\symbdd\bf \zhawi}}
\newcommand\Czhawii{\textcolor{datadriven}{\symbdd\bf \zhawii}}
\newcommand\Cadapt{\textcolor{seqtoseq}{\symbseq\bf \adapt}}
\newcommand\Cdangnt{\textcolor{rules}{\symbrule\bf \dangnt}}
\newcommand\Cforgei{\textcolor{rules}{\symbrule\bf \forgei}}
\newcommand\Cforgeiii{\textcolor{templates}{\symbtempl\bf \forgeiii}}
\newcommand\Cgong{\textcolor{seqtoseq}{\symbseq\bf \gong}}
\newcommand\Charv{\textcolor{seqtoseq}{\symbseq\bf \harv}}
\newcommand\Cnle{\textcolor{seqtoseq}{\symbseq\bf \nle}}
\newcommand\Csheffi{\textcolor{datadriven}{\symbdd\bf \sheffi}}
\newcommand\Csheffii{\textcolor{seqtoseq}{\symbseq\bf \sheffii}}
\newcommand\Cchen{\textcolor{seqtoseq}{\symbseq\bf \chen}}
\newcommand\Cthomsoni{\textcolor{seqtoseq}{\symbseq\bf \thomsoni}}
\newcommand\Cthomsonii{\textcolor{templates}{\symbtempl\bf \thomsonii}}
\newcommand\Ctuda{\textcolor{templates}{\symbtempl\bf \tuda}}
\newcommand\Czhang{\textcolor{seqtoseq}{\symbseq\bf \zhang}}

\makeatletter
\newcolumntype{F}{>{\textfont0=\the@{.}{.}{-1}}c<{\DC@end}}
\newcolumntype{B}{>{\bfseries\textfont0=\the@{.}{.}{-1}}c<{\DC@end}}
\makeatother

\newcommand\mcC[1]{\multicolumn{1}{c}{#1}}
\newcommand\mcB[1]{\multicolumn{1}{B}{#1}}
\newcommand\mcBn[1]{\multicolumn{1}{>{\hspace{-5mm}}B}{#1}}

\usepackage[normalem]{ulem}
\definecolor{darkgreen}{rgb}{0.0, 0.5, 0.0}

\def\ODdel#1{\bgroup\markoverwith{\textcolor{darkgreen}{\rule[0.5ex]{2pt}{1pt}}}\ULon{#1}}

\def\JNdel#1{\bgroup\markoverwith{\textcolor{blue}{\rule[0.5ex]{2pt}{1pt}}}\ULon{#1}}

\def\VRdel#1{\bgroup\markoverwith{\textcolor{magenta}{\rule[0.5ex]{2pt}{1pt}}}\ULon{#1}}

\begin{document}

\title{Evaluating the State-of-the-Art of End-to-End Natural Language Generation: The E2E NLG Challenge}

\author[1,2]{Ondřej Dušek\corref{cor1}}
\ead{odusek@ufal.mff.cuni.cz}

\author[1,3]{Jekaterina Novikova}
\ead{novikova.jekaterina@gmail.com}

\author[1]{Verena Rieser}
\ead{v.t.rieser@hw.ac.uk}

\address[1]{Interaction Lab, Heriot-Watt University, Edinburgh, UK}
\address[2]{Charles University, Faculty of Mathematics and Physics, Prague, Czech Republic}
\address[3]{Winterlight Labs, Toronto, Canada}

\begin{abstract}
This paper provides a comprehensive analysis of the first shared task on End-to-End Natural Language Generation (NLG) and identifies avenues for future research based on the results. 
This shared task aimed to assess whether recent end-to-end 
NLG systems 
can generate more complex output by learning from datasets containing higher lexical richness, syntactic complexity and diverse discourse phenomena.
Introducing novel automatic and human metrics, we compare 62 systems submitted by 17 institutions, covering a wide range of approaches, including machine learning architectures 
-- with the majority implementing sequence-to-sequence models (seq2seq) --
as well as systems based on grammatical rules and templates.
Seq2seq-based systems have demonstrated a great potential for NLG in the challenge. We find that seq2seq systems  
generally score high in terms of word-overlap metrics and human evaluations of naturalness -- with the winning \slug system \citep{juraska_slug2slug:_2018} being seq2seq-based.
However, vanilla seq2seq models often fail to correctly express a given meaning representation if they lack a strong semantic control mechanism applied during decoding.
Moreover, seq2seq models can be outperformed by hand-engineered systems in terms of overall quality, as well as complexity, length and diversity of outputs.
This research has
influenced, inspired and motivated a number of recent studies outwith the original competition, which we also summarise as part of this paper.
\end{abstract}

\maketitle

\section{Introduction}

This paper provides a comprehensive final report and extended analysis of the first shared task on 
End-to-End (E2E) Natural Language Generation (NLG), substantially extending previous reports \citep{novikova2016analogue,Novikova:Sigdial2017,dusek_findings_2018}. 
In addition to this previous work, we provide 
a corrected and extended evaluation of the training dataset, as well as
a detailed discussion of how current state-of-the-art systems address E2E generation challenges, including
semantic accuracy and diversity of outputs, and 
a comparison of techniques used by the submitted systems with systems outside the competition.
We then include a substantially expanded evaluation of
the systems using novel automatic metrics, accounting for output complexity, diversity and semantic correctness. In addition, we provide an analysis of 
system output similarity and confirm that systems using similar techniques, e.g. seq2seq, produce similar outputs.
We also provide a detailed error analysis with examples of system outputs.
This extended evaluation allows us reach some more in-depth insights about the strength and weaknesses of end-to-end generation systems.
Finally, we discuss directions for future work with respect to end-to-end generation, as well as NLG evaluation in general.
In addition, this paper accompanies a release of all the participating systems' outputs on the test set along with the human ratings collected in the evaluation campaign.

Shared challenges have become an established way of pushing research boundaries in the field of Natural Language Processing, with NLG benchmarking tasks running since 2007 \citep{belz:GRE2007}. These previous shared tasks have demonstrated that large-scale, comparative evaluations are vital for identifying future research challenges in NLG \citep{belz:2014}. 
The E2E NLG shared task is novel in that it poses new challenges for recent end-to-end, data-driven NLG systems.
This type of systems promises rapid development of NLG components in new domains by reducing annotation effort:
They jointly learn sentence planning and surface realisation from non-aligned data, e.g. \citep{jurcicek:2015:ACL,wen:emnlp2015,Mei:NAACL2016,Wen:NAACL16,SharmaHSSB16,Dusek:ACL16,vlachos:coling2016}.
As such, these approaches do not require costly semantic alignment between meaning representations (MRs) and the corresponding natural language reference texts (also referred to as ``ground truths" or ``targets"), but they are trained on parallel datasets, which can be collected in sufficient quality and quantity using effective crowdsourcing techniques, e.g.\ \citep{novikova:INLG2016}.

At the start of the E2E NLG Challenge, end-to-end approaches to NLG were limited to small, delexicalised datasets, e.g.\ BAGEL \citep{mairesse:acl2010}, SF Hotels/\hspace{0mm}Restaurants \citep{wen:emnlp2015}, or RoboCup \citep{chen2008learning}.
Therefore, end-to-end methods have not been able to replicate the rich dialogue and discourse phenomena targeted by previous rule-based and statistical  approaches  for language generation in dialogue, e.g.\ \citep{walker2004generation,stent2004trainable,mairesse_personage:_2007,rieser2009natural}.
In this paper, we describe a large-scale shared task based on a new crowdsourced dataset of 50k instances in the restaurant domain (see Section~\ref{sec:data-collection}).
In Section~\ref{sec:dataset}, we show that the dataset poses new challenges, such as open vocabulary, complex syntactic structures and diverse discourse phenomena, and that it inspired multiple extensions and further data collection since its original release.

Our shared task aims to assess whether the novel end-to-end NLG systems are able to produce more complex outputs given a larger and richer training dataset.
We received 62 system submissions by 17 institutions from 11 countries for the E2E NLG Challenge, with about \sfrac{1}{3} of these submissions coming from industry, as summarised in Section~\ref{sec:systems}.
We consider this level of participation an unexpected success, which underlines the timeliness of this task\footnote{Note that, in comparison, the well established Conference in Machine Translation WMT’17 (running since 2006) got 31 institutions submitting to a total of 8 tasks \citep{bojar2017findings}.}
and allows us to reach general conclusions and issue recommendations on the suitability of different methods.

In Section~\ref{sec:challenges}, we analyse how the submitted systems address the challenges posed by the dataset and show that the competition inspired further work on our dataset.
We evaluate the submitted systems by comparing them to a challenging baseline using automatic evaluation metrics (including novel text-based measures) as well as human evaluation (see Section~\ref{sec:eval}).
Note that, while there are other concurrent studies comparing a limited number of end-to-end NLG approaches \citep{Novikova:EMNLP2017,Wiseman:EMNLP17,WebNLG} which emerged during the E2E NLG Challenge, this is the first research to evaluate novel end-to-end generation at scale using human assessment. 

Our results in Section~\ref{sec:results} show a discrepancy between data-driven seq2seq models versus template- and rule-based systems. While seq2seq models generally score high on word-overlap similarity measures and human rankings of naturalness, manually engineered systems score better than some seq2seq systems in terms of overall quality, as well as diversity and complexity of generated outputs.
In Section~\ref{sec:conclusion}, we conclude by laying out challenges for future shared tasks in this area.
We also release a new dataset of 36k system outputs paired with user ratings, which will enable novel research on automatic quality estimation for NLG \citep{specia:MT2010,dusek_referenceless_2017,ueffing_quality_2018,kann_sentence-level_2018,tian_treat_2018}.
All data and scripts associated with the challenge, as well as technical descriptions of participating systems are available at the following URL:
\begin{center}
\url{http://www.macs.hw.ac.uk/InteractionLab/E2E/}
\end{center}

\section{Domain and Task}
\label{sec:domain}

\begin{table}[tb]
\begin{center}
\small
\setlength{\extrarowheight}{3pt}
\begin{tabular}{lll}
\hline
\textbf{Attribute} & \textbf{Data Type} & \textbf{Example value}\\ \hline
\Tstrut name & verbatim string & \it The Eagle, ...\\  
eatType & dictionary & \it restaurant, pub, ...\\  
familyFriendly & boolean & \it Yes / No\\  
priceRange & dictionary & \it cheap, expensive, ...\\  
food & dictionary & \it French, Italian, ...\\  
near & verbatim string & \it market square, Cafe Adriatic, ...\\
area & dictionary & \it riverside, city center, ...\\  
customerRating & enumerable & \it 1 of 5 (low), 4 of 5 (high), ...\\  \hline
\end{tabular}
\end{center}
\caption{Domain ontology of the E2E dataset.}
\label{tab:attr}
\end{table}

\begin{figure}[tb]
\centering\small
\setlength{\extrarowheight}{8pt}
\begin{tabular}{cp{9cm}}
\bf MR & \it name[The Wrestlers], priceRange[cheap], customerRating[1 of 5] \\
\bf reference & The Wrestlers offers competitive prices, but isn't rated highly by customers. \\
\end{tabular}
\vspace{3mm}
\caption{Example pair of an MR and a corresponding human-written reference text.}
\label{fig:pair}
\end{figure}

In general, the task of NLG is to convert an input MR into a natural language utterance consisting of one or more sentences. In this paper, we focus on the case where an end-to-end data-driven generator is trained from simple pairs of MRs and reference texts, without fine-grained alignments between elements of the MR and words or phrases in the reference texts, as in, e.g.\ \citep{jurcicek:2015:ACL,wen:emnlp2015}. An example pair of a MR and a reference text is shown in Figure~\ref{fig:pair}.
We focus on restaurant recommendations in our experiments, which, previously, have been  widely explored in dialogue systems research, e.g.\ \citep{young_hidden_2010,henderson_second_2014,wen_network-based_2017}. However, our E2E dataset is substantially bigger and more complex and than previous NLG training datasets for this domain \citep{mairesse:acl2010,wen:emnlp2015} (see Section~\ref{sec:dataset}), which allows us to assess whether NLG systems are able to learn to produce more varied and complex utterances given enough training examples (cf.~Section~\ref{sec:results}).

For the input representation, we use a format commonly found in task-oriented domain-specific spoken dialogue systems -- unordered sets of \emph{attributes} (slots) and their \emph{values}, e.g.\ \citep{mairesse:acl2010,young_hidden_2010,liu_attention-based_2016}.\footnote{Most dialogue systems also include a general intent of the utterance, such as \emph{inform}, \emph{confirm}, or \emph{request} \citep{young_hidden_2010,wen:emnlp2015,liu_attention-based_2016}. Since our task is focussed on recommendations, this intent would be \emph{recommend/inform} for all our data, and we can therefore disregard it.}
The list of possible attributes used in the MRs in our dataset with example values is shown in Table~\ref{tab:attr}.

\section{Data Collection Procedure}\label{sec:data-collection}

In order to maximise the chances for data-driven end-to-end systems to produce high quality output, we aim to provide training data in sufficient quality and quantity.
We turned to crowdsourcing to collect training data in large enough quantities. We used the CrowdFlower platform\footnote{The CrowdFlower platform was renamed to FigureEight after our study was completed. See \url{https://www.figure-eight.com/}.} to recruit workers.
Previously, crowdsourcing has mainly been used for
evaluation in the NLG community, e.g.\ \citep{Rieser:IEEE14,dethlefs:EMNLP2012}.
However, recent efforts in corpus creation via crowdsourcing have proven to be successful in related tasks.
 For example, \citet{callisonburch:2011} showed that crowdsourcing can result in datasets of comparable quality to those created by professional translators given appropriate quality control methods.
\citet{mairesse:acl2010} demonstrate that crowd workers can produce aligned natural language descriptions from abstract MRs for NLG, a method which also has shown success in related NLP tasks, such as spoken dialogue systems \citep{wang2012crowdsourcing} or semantic parsing \citep{wang-berant-liang:2015:ACL-IJCNLP}.
More recently, data-driven NLG systems, such as \citep{wen_stochastic_2015} and \citep{dusek_context-aware_2016}, have relied on crowdsourcing for collecting training data.

When crowdsourcing corpora for training NLG systems, i.e. eliciting natural language paraphrases for given MRs from workers, the following main challenges arise:
\begin{enumerate}
\item How to ensure the required quality of the collected data?
\item What types of meaning representations can elicit spontaneous, natural and varied data from crowd workers?
\end{enumerate}

In an attempted to address both challenges before collecting the main training dataset for the E2E NLG challenge, we ran a small-scale pre-study published in \citep{novikova:INLG2016}. We briefly summarise the results of this study in this section and apply the successful techniques to the whole data set.

For the pre-study, we prepared a subset of 75 distinct MRs, consisting of three, five or eight attributes from our domain (see Table~\ref{tab:attr}) and their corresponding values in order to evaluate MRs with different complexities.\footnote{The attributes were selected at random, but we excluded MRs that do not contain the attribute \emph{name} as these would not be appropriate for a venue recommendation.}
We then implemented several automatic validation procedures for filtering the crowdsourced data in order to address (1), see Section~\ref{sec:crowdsourcing-validation}.
To address (2), we explored the trade-off between semantic expressiveness of the MR and the quality of crowdsourced utterances elicited for the different semantic representations. In particular, we investigated translating MRs into pictorial representations as used in, e.g. \citep{Williams:2007,black-EtAl:2011:SIGDIAL2011} for evaluating spoken dialogue systems (see Section~\ref{sec:pictorial-mrs}).
In the remainder of this section, we first describe the detailed setup used to crowdsource our data (Section~\ref{sec:crowdsourcing-setup}) and then finally evaluate the pre-study by comparing pictorial MRs to text-based MRs used by previous crowdsourcing work \citep{mairesse:acl2010,wang2012crowdsourcing} in Section~\ref{sec:crowdsourcing-discussion}.

\subsection{Automatic Validation Measures}\label{sec:crowdsourcing-validation}

We used two simple methods to check the quality of crowd workers on CrowdFlower:
First, we only select workers that are likely to be native speakers of English,
following \citet{sprouse2011validation} and \citet{callison2010creating}.
We use IP addresses  to ensure that workers are located in one of three English-speaking countries -- Canada, the United Kingdom, or the United States. In addition, we included a requirement that ``Participants must be native speakers of British or American English" both in the caption of the task listed on CrowdFlower and in the task instructions.
Second, we check whether workers spend at least 20 seconds to complete a page of work. This is a standard CrowdFlower option to control the quality of contributions, and it ensures that the contributor is removed from the job if they complete the task too fast.

We also check the quality of the natural language texts produced by crowd workers for a given MR. In particular, we use three JavaScript validators to ensure that the submitted utterances are well-formed English sentences:

\begin{enumerate}
\item
We check if the ready-to-submit utterance only contains legal characters, i.e.\ letters, numbers and symbols ``\emph{, ' . : ; \pounds}''.

\item We check whether the submitted text is not shorter than the required minimal length, which is an approximation of the total number of characters used for all attribute values in a given MR, as calculated by Eq.~\ref{eq:minimallength}:
\begin{equation}\label{eq:minimallength}
\mbox{\it min. length} = \mbox{\it \# MR characters} -  \mbox{\it \# MR attributes} \times 10
\end{equation}
Here, {\it \# MR characters} is the total number of characters in the given MR; {\it \# MR attributes} is the number of attributes in the given MR; and $10$ is an average length of an attribute name plus two associated square brackets.

\item We check that workers do not submit the same utterance several times.

\end{enumerate}

We ensured by manually checking a small number of initial trial tasks that these automatic validation methods were able to correctly identify and reject 100\% of bad submissions.

\subsection{Meaning Representations: Pictures and Text}\label{sec:pictorial-mrs}

\begin{figure}[tb]
\begin{center}
\small
\begin{tabular}{c>{\raggedright\arraybackslash}m{5cm} >{\centering\arraybackslash}m{5cm}}
1. & \it name[Loch Fyne], eatType[restaurant], familyFriendly[yes], priceRange[cheap], food[Japanese] & \includegraphics[height=3cm]{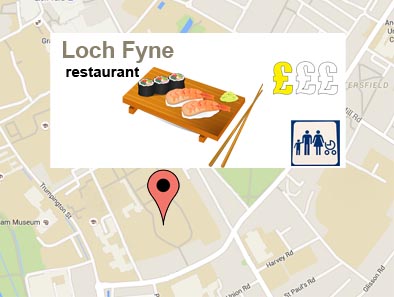} \\
2. & \it name[The Wrestlers], familyFriendly[No], area[riverside], food[Italian], customerRating[5 of 5], priceRange[expensive], \linebreak near[Cafe Adriatic], eatType[restaurant] & \includegraphics[height=3.5cm]{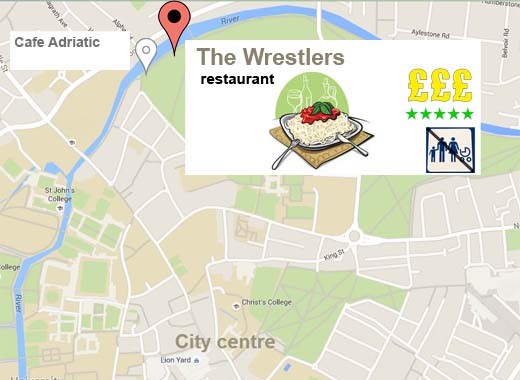} \\
\end{tabular}
\end{center}
\centering
\caption{Examples of  pictorial MRs (left: logical/textual MR, right: corresponding pictorial MR).}
\label{fig:picture}
\end{figure}

In previous crowdsourcing tasks involving MRs, these were typically presented to workers in a textual form of dialogue acts \citep{young_hidden_2010}, such as the following:
\begin{center}
{\em inform(type=hotel, pricerange=expensive)}
\end{center}
However, there is a limit in the semantic complexity that crowd workers can handle when using this type of textual/logical descriptions of dialogue acts~\citep{mairesse:acl2010}. Also, \citet{wang2012crowdsourcing} observed that the chosen semantic formalism  influences the workers' language, i.e.\ crowd workers are primed by the words/tokens and ordering used in the MR.
Therefore, in contrast to previous work \citep{mairesse:acl2010,wen_stochastic_2015,dusek_context-aware_2016}, we explore the usage of different modalities of meaning representation:

\begin{itemize}
\item {\bf Textual/logical MRs} appear as a list of comma-separated attribute-value pairs, where attribute values are shown in square brackets after each attribute (see Figures~\ref{fig:pair} and~\ref{fig:picture}). The order of attributes is randomised so that crowd workers are not primed by the ordering used in the MRs \citep{wang2012crowdsourcing}.

\item {\bf Pictorial MRs} are semi-automatically generated pictures with a combination of icons corresponding to the individual attributes (see Figure~\ref{fig:picture}). The icons are located on a background showing a map of a city, thus allowing to represent the meaning of the attributes \emph{area} and \emph{near}.

\end{itemize}

\subsection{Data Collection Setup}\label{sec:crowdsourcing-setup}

We set up the data collection tasks on the CrowdFlower platform, using the automatic checks described in Section~\ref{sec:crowdsourcing-validation} and using both pictorial and textual MRs as input (see Section~\ref{sec:pictorial-mrs}).
For this pre-study, we collected 1133 distinct utterances from the 75 distinct/unique MRs we prepared. 744 utterances were elicited using the textual MRs, and 498 utterances were elicited using the pictorial MRs. The data collected in the pre-study are freely available for download.\footnote{See \url{https://github.com/jeknov/INLG_16_submission}. The data is not part of the final E2E NLG dataset.}
We later used the same CrowdFlower setup to collect the whole E2E NLG dataset (see Section~\ref{sec:dataset}).

In terms of financial compensation, crowd workers were paid the standard pay on CrowdFlower, which is
 \$0.02 per page (where each page contained 1 MR). Workers  were expected to spend about 20 seconds per page.
 Participants were allowed to complete up to 20 pages, i.e.\ create utterances for up to 20 MRs. 
\citet{mason2010financial} found in their study of financial incentives on Mechanical Turk (counter-intuitively) that increasing the amount of compensation for a particular task does not tend to improve the quality of the results. Furthermore, 
 \citet{callison2010creating}  observed that there can be an inverse relationship between the amount of payment and the quality of work, because it may be more tempting for crowd workers to cheat on high-paying tasks if they do not have the skills to complete them. Following these findings, we did not increase the payment for our task over the standard level.

\subsection{Results and Discussion }\label{sec:crowdsourcing-discussion}

We analysed the collected natural language reference texts, focussing on  textual versus pictorial MRs and their effects on objective measures, such as time taken to collect the data and length of an utterance, and human evaluations of the reference texts collected under the different conditions.
Results in full detail can be found in \citep{novikova:INLG2016}; here we only summarise the main findings. The data analysis showed that:
\begin{itemize}
\item There is no significant difference in the time taken to collect data with pictorial vs.\ textual MRs.
\item The average length of a collected reference text, both in terms of number of characters and number of sentences, depends mainly on the number of attributes associated with the MR, rather than on whether pictures or text were used.
\item Compared to textual MRs, pictorial MRs elicit texts that are significantly less similar to the underlying MR in terms of semantic text similarity \citep{han2013umbc}. We assume that this is because pictorial MRs are less likely to prime the crowd workers in terms of their lexical choices.
\item The human evaluation revealed that reference texts produced from pictorial MRs are rated as significantly ($p<0.01$) more informative  than textual MRs.
Equally, utterances produced from pictorial MRs were considered to be significantly ($p<0.001$) more natural and better phrased than utterances collected with textual MRs.\footnote{Please see \citep{novikova:INLG2016} for a definition of informativeness, naturalness and phrasing.}
\end{itemize}

This shows that pictorial MRs have specific benefits for elicitation of NLG data from crowd workers. This may be because
the lack of priming by lexical tokens in the MRs leads the crowd workers to producing
more spontaneous and natural language, with more variability. As a concrete example of this phenomenon from the collected data, consider the first MR in Figure~\ref{fig:picture}.
The textual version of this MR elicited utterances such as ``{\em Loch Fyne is a family friendly restaurant serving cheap Japanese food.}'' whereas the pictorial MR elicited e.g.\ ``{\em Serving low cost Japanese style cuisine, Loch Fyne caters for everyone, including families with small children.}''

Pictorial stimuli have also been used in other, related NLP tasks, such as crowdsourced evaluations of dialogue systems, e.g.\ \citep{Williams:2007,black-EtAl:2011:SIGDIAL2011}.
\citet{Williams:2007}, for example, used pictures to set dialogue goals for users (e.g.\ to find an  expensive Italian restaurant in the town centre).
However, no analysis  was performed regarding the suitability of such representations.
This experiment therefore has a bearing on the general issue of human natural language responses to pictorial task stimuli, and shows for example that pictorial task presentations can elicit more natural variability in user inputs to a dialogue system.

Of course, there is a limit in the meaning complexity that pictures can express.
We observed that pictorial MRs 
 tend to introduce more noise. In particular,  crowd workers tend to omit information, such as {\em eatType = restaurant}, which is particularly hard to visualise.
 Finally, producing pictorial MRs is a semi-automatic process, which is expensive to run at large scale.

Based on these findings, we decided to use pictorial MRs to collect 20\% of the full dataset and textual MRs for the rest of the data in order to keep noise and production costs low while increasing diversity.
To further increase the data quality and diversity, we collected multiple references per MR to help NLG systems deal with potential noise in the data.

\section{The E2E NLG dataset}\label{sec:dataset}
Using the procedure described in Section~\ref{sec:data-collection},  we crowdsourced a large dataset of 50k instances in the restaurant domain \citep{Novikova:Sigdial2017}. Our dataset is substantially bigger than previous NLG datasets for dialogue in the restaurant domain, i.e.\ BAGEL \citep{mairesse:acl2010} and SF Restaurants (SFRest) \citep{wen:emnlp2015},
which typically only allowed delexicalised data-driven end-to-end approaches (see Section~\ref{sec:dataset-size}).
In addition, we demonstrate that our data is also
more challenging given its lexical richness, syntactic complexity and diverse discourse phenomena.
Following an approach suggested by \citet{Perez-Beltrachini17}, we describe these different dimensions of our dataset and compare them to the BAGEL and SFRest datasets in Sections~\ref{sec:lexical-richness} and~\ref{sec:syntactic-variation}.\footnote{The particular versions of the BAGEL and SFRest datasets used for this research are available from \url{http://farm2.user.srcf.net/research/bagel/} and \url{https://www.repository.cam.ac.uk/handle/1810/251304}, respectively.}

To ensure a fair comparison, we analyse both fully lexicalised and delexicalised versions of all datasets. The lexicalised references in all datasets contained full natural language texts including all restaurant names. This is the default form for the E2E set; small postprocessing steps were taken for the other two sets to achieve a compatible format.\footnote{The BAGEL texts are partially delexicalised by default, so we lexicalised them. SFRest texts were detokenised and adverb/plural markers were postprocessed, e.g. ``restaurant -s'' changed to ``restaurants''.}
To obtain the delexicalised versions, we replaced with placeholders (e.g.\ ``X-slot'') most slot values from open sets that appear verbatim in the data: restaurant names, area names, addresses, and numbers (see Figure~\ref{fig:delex-sets}).\footnote{This included slot values for \emph{name} and \emph{near} in the E2E dataset, \emph{name}, \emph{near}, \emph{phone}, \emph{address}, \emph{postcode}, \emph{count} and \emph{area} in the SFRest dataset, and \emph{name}, \emph{near}, \emph{addr}, \emph{phone}, \emph{postcode} and \emph{area} in the BAGEL set. For BAGEL, the values \emph{citycentre} and \emph{riverside} were excluded from delexicalisation as they do not always appear verbatim in the data.
The delexicalised version of BAGEL is equivalent to how the dataset is distributed by default.
SFRest would allow even more delexicalisation in practice -- food types and price ranges also appear verbatim in the references. We decided to keep these values lexicalised since they are not from open sets and the two other datasets do not allow for easy delexicalisation in this case.\label{fn:delex}}

\begin{figure}[tb]
\scriptsize
\begin{tabular}{l>{\hspace{-3mm}}l>{\hspace{-3mm}\raggedright\arraybackslash}p{10cm}}
\multirow{3}{*}{\bf E2E} & MR & \it name[Green Man], food[French], priceRange[more than £30], area[city centre], familyFriendly[no], near[All Bar One]\\
& Lex. & Green Man is a French restaurant in the city centre. It is not child friendly and is located near All Bar One. It costs more than thirty pounds.\\
& Delex. & \textbf{X-name} is a french restaurant in the city centre . it is not child friendly and is located near \textbf{X-near} . it costs more than thirty pounds .\\
\\
\multirow{3}{*}{\bf SFRest}& MR & \it inform(name=`dosa on fillmore', food=`indian or indpak', address=`1700 fillmore street', phone=4154413672)\\
& Lex. & Dosa on fillmore serves indian and indpak food, the address is 1700 fillmore street, and the phone number is 4154413672. \\
& Delex.& \textbf{X-name} serves indian and indpak food , the address is \textbf{X-address} , and the phone number is \textbf{X-phone} . \\
\\
\multirow{3}{*}{\bf BAGEL}& MR & \it inform(name=``Strada", type=placetoeat, eattype=restaurant, area=citycentre, near=``The Curry House", near=``The Bakers", food=Italian)\\
& Lex. & Strada is an Italian restaurant located near The Curry House and The Bakers in the city centre.\\
& Delex. & \textbf{X-name} is an italian restaurant located near \textbf{X-near} and \textbf{X-near} in the city centre .
\end{tabular}
\caption{Lexicalized and delexicalized examples from all three compared datasets (with slot placeholders highlighted in delexicalized sentences). Note that the dialogue act for E2E is constant (i.e. ``inform") and as such not expressed. SFRest is the only dataset which contains multiple dialogue act types (cf.~the SFRest-inf subset).}
\label{fig:delex-sets}
\end{figure}

Since the E2E and BAGEL datasets contain only restaurant recommendations, i.e.\ cases where the system is providing information (\emph{inform} dialogue acts), whereas SFRest also includes system questions, confirmations, and greetings, we also created a subset of SFRest dubbed SFRest-inf with only \emph{inform} instances for a fairer comparison.

We processed the datasets using the MorphoDiTa part-of-speech tagger \citep{strakova14} to identify tokens, words (as opposed to punctuation tokens) and sentence boundaries. We used the same tagger to preprocess our data for lexical and syntactic complexity analysis.

All code we used for dataset processing and comparison in Sections~\ref{sec:dataset-size}--\ref{sec:syntactic-variation} are freely available for future research under the following URL:
\begin{center}
\url{https://github.com/tuetschek/e2e-stats/}
\end{center}
The main script downloads all three datasets under comparison, installs and patches different third-party metrics tools, and produces the statistics. The same tools are used to compare system outputs in Section~\ref{sec:results-text-metrics}.

\begin{table}[tb]
\begin{center}
\small
\setlength{\extrarowheight}{3pt}
\begin{tabular}{lF>{\hspace{-5mm}}FFF}
\hline
                                             &  \mcC{\bf E2E}      &  \mcC{\bf SFRest}     &  \mcC{\bf SFRest-inf}    &  \mcC{\bf BAGEL} \\
\hline
 Total instances                             &  \mcBn{51,426}      &  5,192                &  3,307                   &  404 \\
 Total MRs                                   &  \mcBn{6,039}       &  1,914                &  1,845                   &  381 \\
 Unique delexicalised MRs                    &  \mcBn{5,963}       &  733                  &  686                     &  156 \\
 Total tokens in all references              &  \mcBn{1,166,000}   &  49,081               &  37,824                  &  6,151 \\
 Total words in all references               &  \mcBn{1,051,093}   &  44,338               &  34,863                  &  5,766 \\
 Total delex. words in all references        &  \mcBn{957,205}     &  37,758               &  28,375                  &  4,671 \\\hdashline[0.5pt/2pt]
 Slots per MR                                &  \mcB{5.74}         &  2.63                 &  2.69                    &  5.48 \\
 \multirow{2}{*}{References per MR}          &  \mcB{8.27}         &  1.91                 &  1.65                    &  1.06 \\
                                             &  \mcC{(1-46)}       &  \mcC{\bf (1-101)}    &  \mcC{(1-33)}            &  \mcC{(1-2)} \\\hdashline[0.5pt/2pt]
 Tokens per reference                        &  \mcB{22.67}        &  9.45                 &  11.44                   &  15.23 \\
 Words per reference                         &  \mcB{20.60}        &  8.54                 &  10.54                   &  14.27 \\
 Delexicalised words per reference           &  \mcB{18.77}        &  7.27                 &  8.58                    &  11.56 \\\hdashline[0.5pt/2pt]
 \multirow{2}{*}{Sentences per reference}    &  \mcB{1.54}         &  1.05                 &  1.07                    &  1.03 \\
                                             &  \mcC{\bf (1-6)}    &  \mcC{(1-4)}          &  \mcC{(1-4)}             &  \mcC{(1-2)} \\
 Tokens per sentence                         &  14.68              &  8.97                 &  10.74                   &  \mcB{14.82} \\
 Words per sentence                          &  13.33              &  8.11                 &  9.90                    &  \mcB{13.89} \\
 Delexicalised words per sentence            &  \mcB{12.15}        &  6.90                 &  8.06                    &  11.26 \\
\hline
\end{tabular}
\end{center}
\caption{Overall size statistics for NLG datasets in the restaurant information domain. All statistics for length of MRs and human references are averages (see Section~\ref{sec:dataset-size} for details). Minimum and maximum numbers of references per MR and sentences per reference are shown in brackets below the average. Highest values on each line are typeset in bold.}
\label{tab:res}
\end{table}

\subsection{Size}
\label{sec:dataset-size}

Table~\ref{tab:res} summarises the main size statistics of all three datasets, plus the inform-only portion of SFRest. 
The E2E dataset is significantly larger than the other sets in terms of the total number of different MRs, the total number of data instances (i.e.\ MR-reference pairs), and especially in terms of the total amount of text in the human references, which is more than 20 times bigger than the next-biggest SFRest.
These differences are even more profound if we consider delexicalisation: almost all MRs in the E2E set are distinct even after delexicalisation, while the number of unique MRs is reduced significantly (by more than half) for the other sets. Delexicalisation also seems to have a less significant effect on the reference texts in the E2E sets than in the other datasets (cf.~the number of delexicalised words vs.~the total number of words).
The high number of instances directly translates to the higher average number of human references per MR, which is 8.27 for the E2E dataset as opposed to less than two for the other sets.\footnote{Note that Refs/MR ratio for the SFRest dataset is skewed: the {\em goodbye()} MR has up to 101 references, but the average is less than 2 references per MR. This is apparent in the SFRest-inf section, which has a much lower maximum number of references.}

While having more data with a higher number of references per MR makes the E2E data more attractive for statistical approaches and enables learning more robust models,
it is also more challenging than previous sets as it contains a larger number of sentences in the human reference texts (up to 6 in our dataset, with an average of 1.54, compared to typically 1--2 for the other sets, which average below 1.1). The sentences themselves are also longer than in the other datasets. This is immediately apparent for SFRest or SFRest-inf, which are up to 40\% shorter in terms of words and tokens. BAGEL's sentences are slightly longer than E2E's on average, but this situation is reversed when the sets are delexicalised.
In addition, the input MRs in the E2E dataset are more complex than in the other sets: the average number of slot-value pairs in our set is twice that of SFRest (even if only the more complex \emph{inform} dialogue acts are considered), and slightly higher than BAGEL.

\begin{table}[tb]
\centering
\small
\setlength{\extrarowheight}{3pt}
\begin{tabular}{lcccc}\hline
\textbf{E2E data part} & \textbf{MRs} & \textbf{References} & \textbf{Slots/MR} & \textbf{Tokens/Ref} \\ \hline
training set & 4,862 & 42,061 & 5.52 & 20.27 \\
development set & \phantom{0,}547 & \phantom{0}4,672 & 6.30 & 24.52 \\
test set & \phantom{0,}630 & \phantom{0}4,693 & 6.91 & 26.76 \\\hdashline[0.5pt/2pt]
full dataset & 6,039 & 51,426 & 5.74 & 22.67 \\\hline
\end{tabular}
\caption{Total number of MRs and human references in the E2E dataset sections and their complexity (average numbers of slots per MR and tokens per reference).}
\label{tab:uniqueMR}
\end{table}

The dataset is split into training, validation and test sets (in a 82-9-9 ratio, see Table~\ref{tab:uniqueMR}).
 We ensure that MRs in our test set are all previously unseen, i.e.\ none of them overlaps with training/development sets, even when restaurant names are removed, unlike the SFRest data \cite[cf.][]{vlachos:coling2016}.
The test set can be considered adversarial since the MRs contained there are somewhat longer/more complex than those in the training set and the references copy this distribution (cf.~Table~\ref{tab:uniqueMR}).

\subsection{Lexical Richness}
\label{sec:lexical-richness}

\begin{table}[tbp]
\centering
\small
\setlength{\extrarowheight}{2pt}
\begin{tabular}{>{\hspace{-2mm}}l<{\hspace{-5mm}}>{\hspace{-5mm}}F>{\hspace{-5mm}}F>{\hspace{-5mm}}F>{\hspace{-5mm}}F}\hline
\bf Lexicalised sets                                             &  \mcC{\bf E2E}      &  \mcC{\bf SFRest}     &  \mcC{\bf SFRest-inf}    &  \mcC{\bf BAGEL} \\
\hline
                  Distinct tokens &            \mcB{2,780} &          1,249 &          1,157 &          601 \\
\multirow{2}{*}{Distinct tokens occurring once} &      \mcB{890}  &    230 &    210 &   205 \\
                                 &      (32\%) &    (18\%) &    (18\%) &   \mcB{(34\%)} \\
                  Distinct lemmas &            \mcB{2,369} &          1,186 &          1,113 &          583 \\
                 Distinct bigrams &           \mcB{30,111} &          5,729 &          4,969 &         1,601 \\
  \multirow{2}{*}{Distinct bigrams occurring once} &   \mcB{13,794}   &   2,582  &   2,272  &   904  \\
  &   (46\%) &   (45\%) &   (46\%) &   \mcB{(56\%)} \\
                Distinct trigrams &          \mcB{100,731} &         11,290 &          9,897 &         2,385 \\
 \multirow{2}{*}{Distinct trigrams occurring once} &    \mcB{56,280}  &   6,832  &   6,091  &  1,667  \\
 &    (56\%) &   (61\%) &   (62\%) &  \mcB{(70\%)} \\\hdashline[0.5pt/2pt]
          Lexical sophistication (LS2) &          0.616 &        0.428 &        0.436 &       \mcB{0.655} \\
                              Type-token ratio (TTR) &          0.002 &        0.027 &        0.032 &       \mcB{0.101} \\
                         Mean segmental TTR (MSTTR-50) &          \mcB{0.706} &        0.648 &        0.626 &       0.654\\\hdashline[0.5pt/2pt]
Unigram entropy & 6.821  & \mcB{7.411}  & 7.375  & 6.773 \\
Bigram entropy & 10.146  & \mcB{10.342}  & 10.202  & 9.043 \\
Trigram entropy & \mcB{12.604}  & 11.830  & 11.766  & 10.159 \\
Bigram next-word conditional entropy\hspace{-1cm} & \mcB{3.213}  & 2.714  & 2.633  & 2.202 \\
Trigram next-word conditional entropy\hspace{-1cm} & \mcB{2.448}  & 1.463  & 1.552  & 1.190\Bstrut \\
\hline
\bf Delexicalised   sets                                          &  \mcC{\bf E2E}      &  \mcC{\bf SFRest}     &  \mcC{\bf SFRest-inf}    &  \mcC{\bf BAGEL} \\
\hline
                  Distinct tokens &            \mcB{2,675} &           504 &           405 &          183 \\
   \multirow{2}{*}{Distinct tokens occurring once} &     \mcB{871}  &    116  &     95  &    56  \\
   &      \mcB{(33\%)} &    (23\%) &     (23\%) &    (31\%) \\
Distinct lemmas & \mcB{2,258} & 437 & 357 & 161 \\
                 Distinct bigrams &          \mcB{26,855} &          3,099 &          2,360 &          659 \\
  \multirow{2}{*}{Distinct bigrams occurring once} &    \mcB{12,379}  &   1,376  &   1,068  &   342  \\
  &    (46\%) &   (44\%) &   (45\%) &  \mcB{(52\%)} \\
                Distinct trigrams &           \mcB{85,736} &          6,383 &          5,033 &         1,129 \\
 \multirow{2}{*}{Distinct trigrams occurring once} &    \mcB{47,881}  &   3,628  &   2,905  &   712  \\
 &    (56\%) &   (57\%) &   (58\%) &   \mcB{(63\%)} \\\hdashline[0.5pt/2pt]
          Lexical sophistication (LS2) &         \mcB{0.600} &        0.323 &        0.317 &       0.317 \\
                              Type-token ratio (TTR) &          0.002 &        0.012 &        0.013 &       \mcB{0.035} \\
                         Mean segmental TTR (MSTTR-50) &          \mcB{0.663} &        0.602 &        0.553 &       0.478 \\\hdashline[0.5pt/2pt]
Unigram entropy & \mcB{6.388}  & 6.305  & 5.944  & 5.294\\
Bigram entropy & \mcB{9.641}  & 9.083  & 8.596  & 7.160\\
Trigram entropy & \mcB{12.122}  & 10.546  & 10.173  & 8.371\\
Bigram next-word conditional entropy\hspace{-1cm} & \mcB{3.140}  & 2.594  & 2.477  & 1.780\\
Trigram next-word conditional entropy\hspace{-1cm} & \mcB{2.446}  & 1.414  & 1.513  & 1.216\\
\hline
\end{tabular}
\caption{Lexical complexity and diversity statistics for NLG datasets in the restarant information domain. Counts for n-grams appearing only once are shown as absolute numbers and proportions of the total number of respective n-grams. Highest values on each line are typeset in bold.}
\label{tab:lexic}
\end{table}

In order to measure various dimensions of lexical richness in the datasets under comparison, we computed statistics on token/unigram, bigram and trigram counts, and we applied the Lexical Complexity Analyser \citep{lu2012relationship},
 as shown in Table~\ref{tab:lexic}.
It is clear that our dataset has a much larger vocabulary -- 2x larger than the second largest SFRest, but more than 5x larger if delexicalised versions of the datasets are considered. This directly translates into the number of distinct lemmas and distinct n-grams; the E2E set has almost 10x more distinct trigrams than SFRest, over 13x more in the delexicalised versions. While the proportion of n-grams only appearing once in the set is slightly lower than in the other datasets, it stays relatively high given the dataset size and narrow domain, and poses a challenging task for end-to-end data-driven approaches.

The traditional measure of lexical diversity is the {type-token ratio} (TTR):
\begin{equation}
\mbox{TTR}(\mbox{text}) = \frac{\#\,\mbox{distinct tokens}}{\#\,\mbox{total tokens}}
\end{equation}
However, it is not a good fit in our case when datasets of different sizes in a narrow domain are compared because the values are inversely proportional to the dataset size.
Therefore, we complement TTR
with the more robust measure of {mean segmental TTR} (MSTTR) \citep{lu2012relationship}, which divides the corpus into successive segments of a given length (50 tokens) and then calculates the average TTR of all segments. The higher the value of MSTTR, the more diverse is the measured text.
Table~\ref{tab:lexic} shows our dataset has higher MSTTR value (0.71) than the other sets ($\leq$0.65).
The difference is even more profound if we consider delexicalised versions of the sets and inform-only MRs in the SFRest data -- 0.66 vs. 0.55 for SFRest-inf and 0.48 for BAGEL.

In addition, we measure {\em lexical sophistication} (LS2) \citep{lu2012relationship}, also known as lexical rareness, 
which is calculated as the proportion of lexical word types not on the list of 2,000 most frequent words generated from the British National Corpus. 
Table~\ref{tab:lexic} shows that while the E2E is more sophisticated than SFRest, it is slightly less so compared to BAGEL.
However, LS2 numbers on the delexicalised sets show that this is mainly caused by lexical slot values --
the delexicalised E2E dataset is almost twice as sophisticated as both SFRest and BAGEL.

Following \citet{oraby_controlling_2018} and \citet{jagfeld_sequence--sequence_2018}, we also use Shannon entropy \cite[p.~61ff.]{manning_foundations_2000} as a measure of lexical diversity in the texts:
\begin{equation}
H(\mbox{text}) = - \sum_{x\,\in\,\mbox{text}} \frac{\mbox{freq}(x)}{\mbox{len}(\mbox{text})} \log_2 \left(\frac{\mbox{freq}(x)}{\mbox{len}(\mbox{text})}\right)
\end{equation}
Here, $x$ stands for all unique tokens/n-grams, \emph{freq} stands for the number of occurrences in the text, and \emph{len} for the total number of tokens/n-grams in the text. We computed entropy over tokens (unigrams), bigrams and trigrams, as shown in Table~\ref{tab:lexic}. We can see that the E2E dataset has slightly lower unigram and bigram entropy than SFRest and higher trigram entropy than any other set. However, when delexicalised, the E2E set shows the highest entropy for any n-gram value. Considering that entropy is a logarithmic measure, the difference is substantial for trigrams -- 12.1 vs.\ the closest 10.5 for SFRest, which amounts to about 2.98$\times$ higher uncertainty.

We further complement Shannon text entropy with n-gram-language-model-style conditional entropy for next-word prediction \cite[p.~63ff.]{manning_foundations_2000}, given one previous word (bigram) or two previous words (trigram):
\begin{equation}
H_{cond}(\mbox{text}) = - \sum_{(c,w)\,\in\,\mbox{text}} \frac{\mbox{freq}(c,w)}{\mbox{len}(\mbox{text})} \log_2 \left(\frac{\mbox{freq}(c,w)}{\mbox{freq}(c)}\right)
\end{equation}
Here, $(c,w)$ stands for all unique n-grams in the text, composed of $c$ (context, all tokens but the last one) and $w$ (the last token).
Conditional next-word entropy gives an additional, novel measure of diversity and repetitiveness: The more diverse a text is, the less predictable is the next word given previous word(s); on the other hand, the more repetitive the text, the more predictable is the next word given previous word(s). The values for all the datasets are again shown in Table~\ref{tab:lexic}, and they demonstrate clearly that E2E data is much more diverse than SFRest or BAGEL. Note also that lexicalisation has a much smaller effect on this measure. In the delexicalised version, the difference against the closest SFRest (2.446 vs.\ 1.414) indicates about 2.04$\times$ more uncertainty on next-word prediction given two previous words.

\begin{figure}[tb]
\centering
\includegraphics[width=12cm]{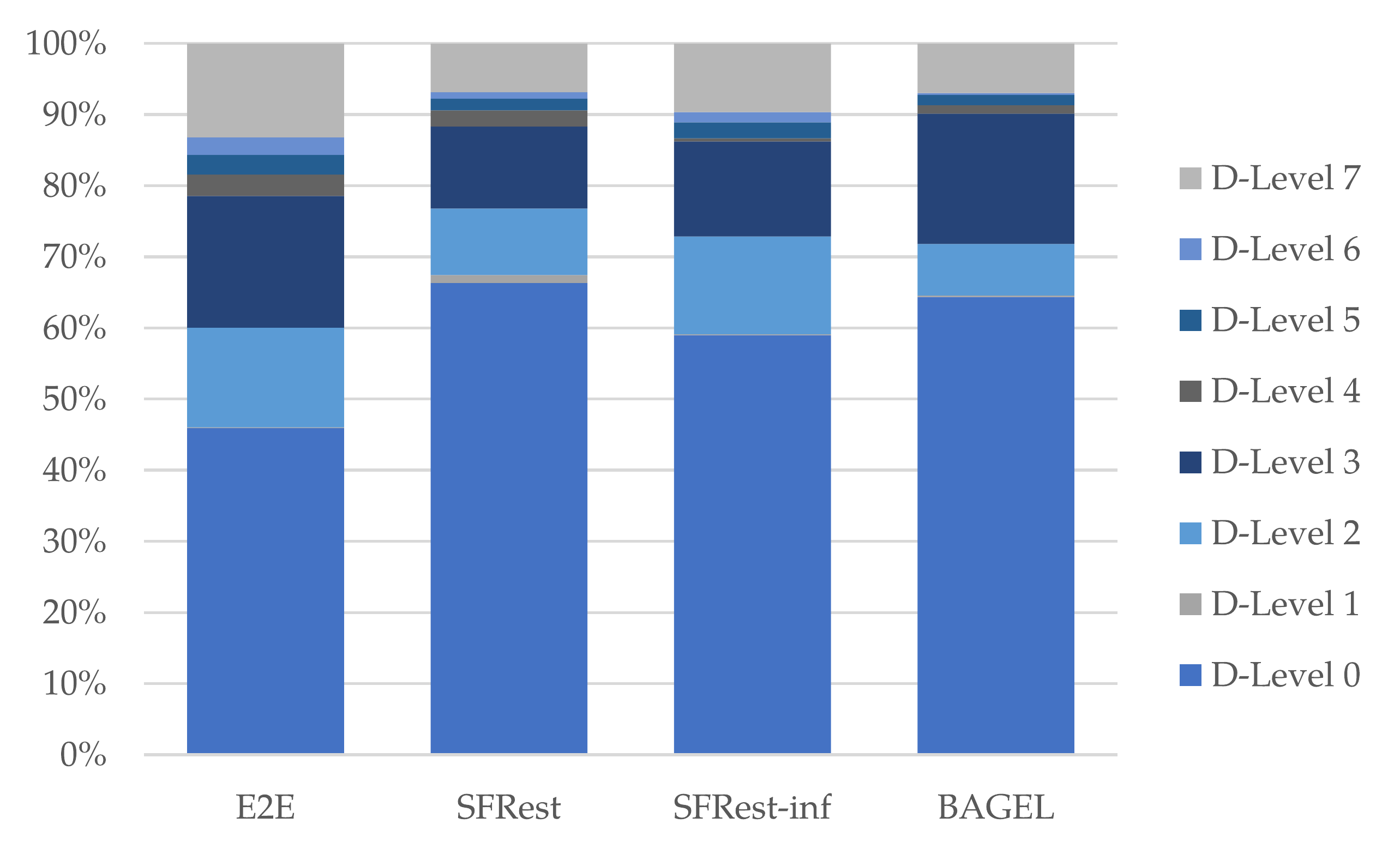}
\caption{
D-Level sentence distribution of the datasets under comparison. 
}
\label{fig:dlevels}
\end{figure}

\subsection{Syntactic Complexity}
\label{sec:syntactic-variation}

We used the D-Level Analyser \citep{lu2009automatic} 
to evaluate the syntactic complexity of human references in our data using the revised D-Level Scale \citep{covington_how_2006}.
We used the syntactic constituency parser of \citet{collins_three_1997} to preprocess the sentences for the D-Level Analyser.\footnote{We used the Model 2 variant of the parser as instructed by the D-Level Analyser website at \url{http://www.personal.psu.edu/xxl13/downloads/d-level.html}.}
The D-Level scale has eight levels of syntactic complexity, where levels 0 and 1 include simple or incomplete sentences and higher levels include sentences with more complex structures, e.g.\ sentences joined by a subordinating conjunction, more than one level of embedding etc.
Figure~\ref{fig:dlevels} shows the D-Level distribution in all three datasets.

The largest proportion of the datasets is composed of simple sentences (levels~0 and~1), but the proportion of simple texts is much lower for the E2E NLG dataset (46\%) compared to others (59-66\%). Examples of simple sentences in our dataset include: ``The Vaults is an Indian restaurant'', or ``The Loch Fyne is a moderate priced family restaurant''.

The majority of our data, however, contains more complex, varied syntactic structures, including phenomena explicitly modelled by early statistical approaches to NLG \citep{stent2004trainable,walker2004generation}.\footnote{Some of the systems in the competition as well as multiple follow-up works are specifically taking advantage of the added complexity in our dataset to produce more varied outputs (see Sections~\ref{sec:data-augmentation} and~\ref{sec:systems-outside}).}
For example, clauses may be joined by a coordinating conjunction (level~2), e.g. ``Cocum is a very expensive restaurant \textit{but} the quality is great''.
There are 14\% level-2 sentences in the E2E dataset; BAGEL only has 7\% and SFRest 9\%, but \emph{inform} MRs in SFRest contain a similar proportion as our set.
Level~3 sentences in our domain are mainly those with object-modifying relative clauses, e.g.\ ``There is a pub called Strada \emph{which} serves Italian food.''
The E2E dataset contains 18\% level-3 sentences, similar to BAGEL but more than SFRest's 12\% (13\% in \emph{inform} MRs).
The levels 4-5 are not very frequent in any of the datasets.
Sentences may contain verbal gerund (\emph{-ing}) phrases (level~4), either in addition to previously discussed structures or separately, e.g.\ ``The coffee shop Wildwood has fairly priced food, \textit{while being} in the same vicinity as the Ranch'' or ``The Vaults is a family-friendly restaurant \textit{offering} fast food at moderate prices''.
Subordinate clauses are marked as level~5, e.g. ``\textit{If} you like Japanese food, try the Vaults''.

The highest levels of syntactic complexity involve sentences containing referring expressions (``The Golden Curry provides Chinese food in the high price range. \textit{It} is near the Bakers''), non-finite clauses in adjunct position (``\textit{Serving} cheap English food, as well as \textit{having} a coffee shop, the Golden Palace has an average customer rating and is located along the riverside'') or sentences with multiple embedded structures from previous levels. As Figure~\ref{fig:dlevels} shows, our dataset has a substantially higher proportion of level-6-7 sentences -- 15\%, compared to 7\% for BAGEL and 8\% for SFRest (11\% in \emph{inform} MRs).

On average, sentences in the E2E dataset are much more syntactically complex than in the other datasets under comparison: the mean D-Level for E2E data is 2.17, compared to BAGEL's 1.32 and SFRest's 1.25 (1.57 for \emph{inform}-only MRs).

\subsection{Attribute Coverage}\label{sec:content-selection}

\begin{table}[tb]
\begin{center}
\small
\setlength{\extrarowheight}{3pt}
\begin{tabular}{lccc}
\hline
& \bf E2E & \bf SFRest & \bf BAGEL \\
\hline
Fully covered & 30  & 47 & 50 \\
Missing content & 11 & \phantom{0}0 & \phantom{0}0 \\
Additional content & \phantom{0}9 & \phantom{0}3 & \phantom{0}0 \\\hline
\end{tabular}
\end{center}
\caption{Coverage of MR attributes in references as measured manually on a random sample of 50 MR-reference pairs for each dataset. The numbers indicate the absolute number of instances falling into the given category, out of 50.}

\label{tab:semantics}
\end{table}

Our crowd workers  were asked to verbalise all 
information from the MR; however, they were not penalised if they skip an attribute (cf.\ Section~\ref{sec:crowdsourcing-discussion}).
This feature makes generating text from our dataset more challenging as the NLG systems need to deal with a certain amount of noise, i.e.\ attributes not being verbalised in the human reference texts.
In order to measure the extent of
this phenomenon, we examined a random sample of 50 MR-reference pairs in all three datasets under comparison.
An MR-reference pair was considered ``fully covered'' if all attribute values present in the MR are verbalised in the reference. It was marked
as ``additional content'' if the reference contains information not present in the MR, and as ``missing content'' if the MR contains information not present in the reference.

The results of our sample probe in Table~\ref{tab:semantics} indicate that roughly 40\% of our data contains either additional or omitted information. 
In order to help NLG systems account for this variation, we collected multiple references per MR (also see Table~\ref{tab:res}).

This variation often concerns the attribute-value pair \emph{eatType=restaurant}, which is either omitted (``Loch Fyne provides French food near The Rice Boat. It is located in riverside and has a low customer rating'') or added in case \emph{eatType} is absent from the MR (``Loch Fyne is a low-rating riverside French \underline{restaurant} near The Rice Boat'').\footnote{Note that 
 inclusion of this attribute is mainly due to historical reasons, following SFRest and BAGEL. }
As discussed in Section \ref{sec:crowdsourcing-discussion}, pictorial MRs might be a possible source of this phenomenon where {\em eatType=restaurant, eatType=pub}, etc.\ is difficult to illustrate.

\subsection{Following-on Datasets}
\label{sec:inspired-datasets}

Since the E2E dataset was first published in \cite{Novikova:Sigdial2017}, it inspired multiple extensions (cf.\ also Section~\ref{sec:systems-outside}). 
The work of \citet{juraska_characterizing_2018} adds further automatic annotation of contrast and emphasis to study the style of generation outputs. 
\citet{oraby_controlling_2018,oraby_neural_2018} used E2E MRs in combination with the Personage generator \cite{mairesse_personage:_2007,mairesse_controlling_2011} to create a synthetic corpus to examine how neural models can learn various stylistic properties. 
\citet{reed_can_2018} then combine Personage-generated data for E2E MRs with the original crowdsourced data and add more supervision specifically to study how neural generators perform various sentence planning operations (sentence aggregation, distribution of content and discourse relations).
\citet{balakrishnan_constrained_2019} enhance both MRs and references in the E2E set: their enhanced tree-structured MRs include explicit fine-grained dialogue acts (inform, contrast, recommend) automatically obtained from the corresponding references; the enhanced references mark explicitly these dialogue acts as well as which phrase corresponds to which MR attribute.
\citet{roberti_copy_2019} present an enhancement of the E2E of a different kind -- they 
include many more restaurant names and food types than present in the original to make the open vocabulary problem more apparent and study it in detail.
Finally, the recent Yelp restaurant information dataset of \citet{oraby_curate_2019} is also inspired by E2E, but takes a different approach -- collecting large-scale web-based data with automatic annotation.

\section{Systems in the Competition}\label{sec:systems}

\begin{table}[tp]
\begin{center}
\scriptsize
\setlength{\extrarowheight}{3pt}
\begin{tabular}{>{\raggedright}p{2.5cm}>{\raggedright}p{1.5cm}>{\raggedright}p{1.2cm}<{\hspace{-3mm}}l>{\raggedright}p{2.3cm}>{\raggedright\arraybackslash}p{2cm}}
\hline
\bf System   & \bf Architecture   & \bf Delex.\ slots   & \hspace{-2mm}\bf Copy\hspace{-2mm}  & \bf Semantic control   & \bf Data augmentation / diversity\\\hline

\Ctgen\ \citep{Novikova:Sigdial2017}  & seq2seq (\tgen)   & \it name, near   &   & MR classification reranking   & \\\hdashline[0.5pt/2pt]
\Cadapt\\\citep{elder_e2e_2018}  & seq2seq \mbox{(OpenNMT-py)}   & none    & \checkmark & none   &   enriching MR by  output words\\
\Cchen~\citep{chen_general_2018}   & seq2seq   & none    & \checkmark  & attention memory   &    \\
\Cgong~\citep{gong_technical_2018}  & seq2seq (\tgen)   & \it name, near    &   & MR classification reranking   &    \\
\Charv\\\citep{gehrmann_end--end_2018}  & seq2seq   & none    & \checkmark  & coverage penalty reranking   & diverse ensembling \\
\Cnle~\citep{agarwal_char-based_2018}   & char seq2seq (tf-seq2seq)   & none    &   & MR classification reranking   &  \\
\Csheffii~\citep{chen_shefeld_2018}  & seq2seq   & \it name, near    &   & none    &   \\
\Cslug\\\citep{juraska_slug2slug:_2018}  & seq2seq   & \it name, near   &   & slot aligner reranking   & using sub-MRs  and aligned sentences \\
\Cslugalt\\\citep{juraska_slug2slug:_2018}  \\ \emph{(late submission)} & seq2seq & \it name, near   &   & slot aligner reranking   & using only complex training sentences \\
\Ctntnlgi\\\citep{oraby_tntnlg-personage_2018} & seq2seq (\tgen)   & \it name, near   &   & MR classification reranking   & using Personage \\
\Ctntnlgii\\\citep{tandon_tntnlg-mr_shuffle_2018} & seq2seq (\tgen)   & \it name, near   &   & MR classification reranking   & shuffling MRs \\
\Cthomsoni\\\citep{schilder_e2e_2018}   & seq2seq (tf-seq2seq)   & \it name, near,\\ priceRange, \mbox{customerRating}  &   & none    &  \\
\Czhang\\\citep{zhang_attention_2018}   & sub-word seq2seq  & none    &   & attention regularisation   &    \\\hdashline[0.5pt/2pt]
\Csheffi~\citep{chen_shefeld_2018}   & linear classifiers \\ + LOLS   & \it name, near   &   & 2-step prediction with slots   & using only references with highest average word frequency   \\
\Czhawi~\citep{deriu_end--end_2018} & RNN language model   & \it name, near   &   & SC-LSTM (semantic gates), MR classification loss + reranking   &    first word control\\
\Czhawii~\citep{deriu_end--end_2018}  & RNN language model   & \it name, near    &   & SC-LSTM \\ (semantic gates)   &   first word control\\\hdashline[0.5pt/2pt]
\Cdangnt~\citep{nguyen_structure-based_2018}  & rule-based   & all    &   & implied by architecture   &  \\
\Cforgei~\citep{mille_forge_2018}  & grammar   & all   &   & implied by architecture   &    \\\hdashline[0.5pt/2pt]
\Cforgeiii~\citep{mille_forge_2018}  & templates   & all    &   & implied by architecture    & \\
\Cthomsonii\\\citep{schilder_e2e_2018}   & templates   & all    &   & implied by architecture   & \\
\Ctuda~\citep{puzikov_e2e_2018}   & templates   & all    &   & implied by architecture    & \\\hline
\end{tabular}
\end{center}
\caption{A full list of the primary systems participating in the E2E challenge, with their basic architecture and other properties (list of delexicalised slots, presence of a copy mechanism, control of semantic MR coverage on the output, data augmentation and output diversity techniques). System architectures are coded with colours and symbols: \textcolor{seqtoseq}{\symbseq seq2seq}, \textcolor{datadriven}{\symbdd other data-driven}, \textcolor{rules}{\symbrule rule-based}, \textcolor{templates}{\symbtempl template-based}.}\label{tab:systems}
\end{table}

The initial idea of the E2E NLG Challenge was first presented in \citep{novikova2016analogue}. The interest and active participation in the E2E Challenge has by far outperformed our expectations. We received a total of 62 submitted systems by 17 institutions from 11 countries, with about \sfrac{1}{3} of these submissions coming from industry.
In accordance with ethical considerations for NLP shared tasks \citep{ethicalSharedTasks}, we allowed researchers to withdraw or anonymise their results after obtaining automatic evaluation metrics results (cf.\ Section~\ref{sec:automatic-metrics}).
Two groups from industry withdrew their submissions and one group asked to be anonymised
after obtaining automatic evaluation results.
A full list of all the remaining submissions is given in Table~\ref{tab:all-systems-wbms} in the Appendix (including their automatic metric scores).

We asked each participating team to identify 1-2 primary systems, which resulted in 20 systems by 14 groups.
Each primary system is described in a short technical paper (available on the E2E NLG Challenge website)\footnote{\url{http://www.macs.hw.ac.uk/InteractionLab/E2E/}} and was
 evaluated both by automatic metrics and human judges (see Section~\ref{sec:eval}).
We compare the primary systems to a baseline system we provided ourselves (see Section~\ref{sec:baseline}).
A detailed overview of all the primary systems is given in Table~\ref{tab:systems}.
In the following, we describe 
the systems in terms of different architectures; see Sections~\ref{sec:seq2seq-based}--\ref{sec:template-based}.

\subsection{Baseline System}\label{sec:baseline}

\begin{table}[tbp]
\begin{center}
\small
\setlength{\extrarowheight}{3pt}
\begin{tabular}{lccccc}\hline
& \bf \bleu & \bf \nist & \bf \meteor & \bf \rouge & \bf \cider \\\hline
\tgen (development set) & 0.6925 & 8.4781 & 0.4703 & 0.7257 & 2.3987 \\\hline
\end{tabular}
\end{center}
\caption{\tgen performance on the development set (see Section~\ref{sec:automatic-metrics} for a description of the metrics).}\label{tab:dev-tgen}
\end{table}

To establish a baseline on the task data, we use \tgen \citep{Dusek:ACL16}.\footnote{\tgen is freely available at \url{https://github.com/UFAL-DSG/tgen}.} \tgen is based on the sequence-to-sequence model with attention (seq2seq) \citep{bahdanau_neural_2015}, an encoder-decoder recurrent neural network (RNN) architecture.
In addition to the standard seq2seq model with LSTM cells \citep{hochreiter_long_1997}, \tgen uses beam search for decoding and an LSTM-based
reranker over the top $k$ outputs, penalising those outputs that do not verbalise all attributes from the input MR.
\tgen was previously tested on the BAGEL and SFRest datasets, where it reached state-of-the-art performance \cite[p.~88ff.]{dusek_novel_2017}.

As \tgen does not handle unknown vocabulary well,
the sparsely occurring string attributes (see Table~\ref{tab:attr}) \emph{name} and \emph{near} are delexicalised (see Section~\ref{sec:delexicalisation}). The main seq2seq model is trained by minimising cross entropy using the Adam algorithm \citep{kingma_adam:_2015} in direct token-by-token generation of surface strings; the reranker is trained to detect the presence of all attributes from the input MR.\footnote{We use a learning rate of 0.0005, cell size 50, batch size 20, beam size 10, maximum encoder and decoder lengths 10 and 80, respectively, and up to 20 passes through training data with early stopping. The reranker uses the same parameters, except for a higher learning rate (0.001). See
\citep{Novikova:Sigdial2017} for more details.}
Based on evaluation on the development part of the E2E dataset using automatic metrics (see Table~\ref{tab:dev-tgen}), as well as manual cursory checks, \tgen appears to be a strong baseline, capable of generating fluent and relevant outputs in most cases.

\subsection{Seq2seq-based systems}\label{sec:seq2seq-based}

Systems based on the popular sequence-to-sequence architecture \citep{sutskever_sequence_2014,bahdanau_neural_2015} represent the biggest group of systems participating in the challenge (12 out of 20 primary systems).
All the seq2seq-based systems use beam search,
and most of them further enhance the basic seq2seq architecture in a number of ways. 

Several systems are built on top of previous systems and toolkits.
A number of systems are based on the \tgen baseline and aiming to improve it:
\tntnlgi \citep{oraby_tntnlg-personage_2018} and \tntnlgii \citep{tandon_tntnlg-mr_shuffle_2018} are using \tgen with two different data augmentation techniques (see Section~\ref{sec:data-augmentation}).
\gong \citep{gong_technical_2018} trains \tgen with fine-tuning by the REINFORCE algorithm \citep{williams_simple_1992}.
Two systems are based on the tf-seq2seq toolkit \citep{britz_massive_2017}:
\nle \citep{agarwal_char-based_2018} built a character-to-character seq2seq (using simply characters of the original MR as inputs), \thomsoni \citep{schilder_e2e_2018} use a regular word-based model.
The \adapt system \citep{elder_e2e_2018} is based on OpenNMT-py \citep{klein_opennmt:_2017}. It uses pointer networks (a form of a copy mechanism \citep{vinyals_pointer_2015}) and a two-step generation where the first step enriches the input MR for diversity (see Section~\ref{sec:data-augmentation}).

Several other systems use custom seq2seq implementations.
\slug and \slugalt \citep{juraska_slug2slug:_2018} use an ensemble of two bidirectional LSTM encoders and one convolutional encoder, all paired with an attention LSTM decoder (incl.\ self-attention).
\harv \citep{gehrmann_end--end_2018} use a seq2seq model with multiple additions for MR coverage and diversity (see Sections~\ref{sec:semantic-control} and~\ref{sec:data-augmentation}). 
\sheffii's model \citep{chen_shefeld_2018}, on the other hand, is a vanilla seq2seq setup with LSTM cells.
\chen \citep{chen_general_2018} presents a seq2seq model with a custom-tailored input data representation: 2-part input embeddings, which divide into slot name and value token embeddings.
\zhang \citep{zhang_attention_2018} apply a seq2seq model with CAEncoder \citep{zhang_context-aware_2017}, which adds a second layer over a bidirectional encoder with GRU cells \citep{cho_learning_2014}, summarising both directional encoders.

\subsection{Other data-driven systems}\label{sec:non-seq2seq-datadriven}

Two groups submitted fully trainable systems that are not based on the seq2seq architecture. First, \zhawi and \zhawii \citep{deriu_end--end_2018} use an RNN language model with semantically conditioned LSTM (SC-LSTM) cells \citep{wen:emnlp2015} and a 1-hot encoding of input MR slot values. The two system variants differ in the presence of an additional semantic control mechanism (see Section~\ref{sec:semantic-control}).

\sheffi \citep{chen_shefeld_2018} is the only non-neural fully data-driven system submitted to the challenge. It is based on imitation learning using
 linear classifiers \citep{crammer_adaptive_2009} in a two-level generation approach, where the classifiers first select the next slot to be realised 
 and then the corresponding word-by-word realisation of that slot \citep{vlachos:coling2016}.
 The classifiers are trained using the Locally Optimal Learning to Search (LOLS) imitation learning framework \citep{chang_learning_2015}, optimising for \bleu, \rouge, and slot error (cf.\ Section~\ref{sec:automatic-metrics}).

\subsection{Rule-based systems}\label{sec:rule-based}

There are two rule-based entries in the E2E challenge:
First, the \dangnt system \citep{nguyen_structure-based_2018} uses a two-step rule-based setup, where the first step determines the appropriate phrases to use for a delexicalised sentence; the second step selects the appropriate phrases to lexicalise slot values. 
Second, the \forgei system \citep{mille_forge_2018} is a rule-based pipeline using grammars based on the Meaning-Text Theory \citep{melcuk_dependency_1988}. It matches the MR to handcrafted per-slot semantic templates, applies aggregation rules to build sentences, and realises the aggregated sentence structures into surface text.

\subsection{Template-based systems}\label{sec:template-based}

Three entries in the E2E challenge are based on traditional template filling.
\forgeiii \citep{mille_forge_2018} and \thomsonii \citep{schilder_e2e_2018} take a very similar approach: They mine templates from data by delexicalising slot values.
\tuda \citep{puzikov_e2e_2018}, on the other hand, uses templates manually designed by the system authors; the templates are not based on the dataset directly, they are only informed by the data.

\section{Addressing the Challenges}\label{sec:challenges}

In this section, we focus on how the competing primary systems address specific challenges posed by the task: vocabulary unseen in training (Section~\ref{sec:delexicalisation}), control of semantic coverage of the input MR (Section~\ref{sec:semantic-control}), and producing diverse outputs (Section~\ref{sec:data-augmentation}). We also include an overview of alternative approaches to addressing these challenges in Section~\ref{sec:systems-outside}.

\subsection{Open Vocabulary}\label{sec:delexicalisation}

All systems in the challenge have a way of addressing the open vocabulary in the data.
In closed-domain setups, slot values are the usually the only part of data where open vocabulary is present, as e.g.\ is the case of the \emph{name} and \emph{near} slots in our dataset (see Table~\ref{tab:attr}). 
The common approach to dealing with open vocabulary in NLG systems is to use delexicalisation (\citealp{wen:emnlp2015}; see also Section~\ref{sec:dataset}), i.e.\ replacing slot values with placeholders during training and generation time (both in input MRs and training sentences). This approach is indeed one of the principles of template-based systems; accordingly, all template-based entries in the E2E Challenge use full delexicalisation of all slot values (except, perhaps, the binary-valued \emph{familyFriendly}; cf.~Table~\ref{tab:systems}). Both rule-based systems also perform full delexicalisation.

The data-driven systems submitted to our challenge mostly opt for partial delexicalisation (see Table~\ref{tab:systems}); the prevailing approach is to delexicalise only the values of the \emph{name} and \emph{near} slots, which allows for very simple pre- and postprocessing since these values usually appear verbatim in the outputs.\footnote{Unlike other slot values, e.g., \emph{area=riverside} might appear as ``near the river''. Cf.\ also our remarks on delexicalisation in Section~\ref{sec:dataset} and Footnote~\ref{fn:delex}.} \thomsoni is the only data-driven system to use a stronger delexicalisation, which also includes the \emph{priceRange} and \emph{customerRating} slots. \slug and \slugalt are the only systems to treat values with different morpho-syntactic properties differently (e.g., a value requiring ``an'' instead of ``a'' as an article).

Five of the seq2seq systems in the challenge opted for using no delexicalisation and employ alternative ways of addressing open vocabulary: \adapt, \chen and \harv use a copy mechanism (cf.~Section~\ref{sec:seq2seq-based}), which allows the system to copy some of the tokens from the input instead of generating them anew.
\zhang operates over sub-word units instead of words; these are determined by the byte-pair encoding algorithm and can combine to create previously unseen words \citep{sennrich_neural_2016}.
\nle's seq2seq system operates on the character level.

\subsection{Semantic Control}\label{sec:semantic-control}

Most of the participating systems explicitly attempt to realise all slots and thus cope with the noise in the training data (cf.\ Section~\ref{sec:content-selection}). Full realisation is implied for template and rule-based systems as the templates and rules always relate to specific slots and are chosen based on the slots in the input MR. On the other hand, vanilla seq2seq systems have no way of controlling whether all input slots have been realised. While attention models \citep{bahdanau_neural_2015} certainly have an influence on this, they are not explicitly trained to attend exactly once to each slot in a vanilla seq2seq setup.
Therefore, most seq2seq systems include an additional tool checking the realised parts of the input MR on the output (cf.~Table~\ref{tab:systems}).

The most frequent approach among the E2E submissions is a MR classification reranker \citep{Dusek:ACL16}. Here, the generator first produces multiple outputs using beam search, then these are tested for the presence of all slots from the input MR, and deviations from the input are penalised. Apart from the \tgen baseline (using a RNN MR classifier, see Section~\ref{sec:baseline}), this approach is also taken by all systems based on \tgen (\tntnlgi, \tntnlgii, \gong) as well as \nle, which uses a logistic regression classifier. \slug and \slugalt apply a very similar approach: they use a heuristic slot aligner (trained on words and phrases from training data and WordNet) to align outputs to the input MR and penalise for any unaligned slots. \harv do not build a separate classifier or aligner, but use the sum of weights from the attention model (which should not exceed 1 for each token of the input MR) in a penalty term for reranking.

Two seq2seq systems use a direct modification of the attention mechanism instead of reranking at decoding time. \chen includes attention memory (sum of attention distributions so far in the generation process) as an additional input to the attention model. \zhang adds an attention regularisation loss term to the training process, which attempts to keep the sum of weights close to 1 for each input MR token, similarly to \harv's penalty term. Three systems, \adapt, \thomsoni and \sheffii, do not use any explicit semantic control mechanism.

The non-seq2seq data-driven systems use specific mechanisms to maintain input MR coverage. \zhawi and \zhawii are based on SC-LSTM cells \citep{wen:emnlp2015}, which include a special gate that keeps track of slots covered so far in the MR. In addition, \zhawi uses convolutional MR classifiers to rerank beam search outputs similarly to most seq2seq systems; however, this classification is also used in an additional loss term during training. The \sheffi system explicitly decides which slot to verbalise next using a separate slot-level classifier, which is optimised to cover the input MR.

\subsection{Data Augmentation and Diversity}\label{sec:data-augmentation}

The design of the E2E dataset attempts to provide higher text diversity (see Section \ref{sec:dataset}), and several challenge participants made use of this. Others modified the training set simply to achieve better output quality.

Several systems aim at higher output quality by using data augmentation. \tntnlgi enriches input MRs by prepending
 them with the corresponding outputs of the Personage generator \citep{mairesse_personage:_2007}, with the aim to generate
more diverse output. \tntnlgii aims to boost the robustness of the baseline \tgen system by re-shuffling slots in the input MRs.
\slug uses single sentences from the training data with corresponding aligned parts of the original MR. This increases the amount of training data available and simplifies the task by breaking outputs into smaller (partially) aligned units. \slugalt, on the other hand, only uses training instances involving complex sentences in an attempt to provide more sophisticated outputs.
On the other hand, the system of \sheffi is trained using only one reference text per training MR; the reference text with the highest average word frequency is selected. While this approach is likely to decrease
output diversity, the authors use it to stabilise system training.
\harv takes yet another approach in order to both stabilise training and increase diversity, called diverse ensembling \citep{guzman-rivera_multiple_2012}. In an expectation-maximisation fashion, they split the training data instances into subsets that exhibit similar structural properties and style in the natural language references,
then train different models on these subsets and deploy them as an ensemble.

Two teams attempt to increase output diversity by directly modifying the generation process.
The \zhawi and \zhawii systems use a first word control mechanism: they generate outputs starting with all (frequent enough) first words from the training set, then select the final output by sampling.
\zhawi only samples among semantically correct outputs (see Section~\ref{sec:semantic-control}).
\adapt takes a different approach, adding a preprocessing step before the main generator, which decides upon specific words that should appear on the output. These are then used to enrich the input MR in the main generation step, providing more diversity on the input.

\subsection{Systems outside the competition and E2E-inspired work}
\label{sec:systems-outside}

Solving the challenges outlined above is an ongoing effort addressed by many recent systems. Here we briefly summarise other attempts by systems outside the competition for completeness. Note that many of these approaches are very recent and have been published only after the E2E NLG Challenge ended; some them are even inspired by the challenge and work with the E2E dataset.

Apart from delexicalisation, which is most often used in the E2E NLG Challenge, various variants of the copy mechanism are the most prominent approach to address open vocabulary in NLG \citep{Wiseman:EMNLP17,lebret_neural_2016,bao_table--text:_2018,kaffee_learning_2018,wang_describing_2018}. Among works using the E2E dataset, \citet{shimorina_handling_2018} combine a copy mechanism with delexicalisation. In contrast, \citet{freitag_unsupervised_2018} use subwords and recast the NLG model as a denoising autoencoder, with shared input and output embeddings (starting from slot values and ``filling in'' the rest of the sentence on the output). \citet{roberti_copy_2019} explore the use of character-based models and extend the E2E dataset to include a wider variety of restaurant names to showcase their approach.

Attempts at improving semantic accuracy of the generated texts show a wider variety of approaches. \citet{kiddon_globally_2016} use a ``checklist model'' -- the decoder keeps a vector of items used so far during the generation; this is similar to semantic gates of \citet{wen:emnlp2015}, which have been used by the \zhawi and \zhawii systems in our challenge (see Section~\ref{sec:semantic-control}). \citet{tran_neural-based_2017} use a two-level attention model (composed of a standard attention model and a ``refiner'', an attention-over-attention module) to improve semantic coverage. \citet{nema_generating_2018} combine semantic gating and two-level attention (with attention over slots, slot values, and a combination thereof).
The system of \citet{su_natural_2018} and \citet{su_investigating_2018}, which is developed on E2E data, explores using multi-level decoder, adding linguistic complexity gradually to maintain output integrity; this is in fact similar to \citet{freitag_unsupervised_2018}'s approach. A follow-up by \citet{su_dual_2019} explores using multi-objective optimisation for both NLG and language understanding, where the latter serves as regularisation for the former.
Other authors working on the E2E dataset explore supplementary inputs for improving semantic correctness: \citet{reed_can_2018} use an additional supervision signal indicating the desired number of sentences to generate, \citet{freitag_unsupervised_2018} show that additional unlabelled training data improves semantic coverage in their denoising-autoencoder-based NLG model and \citet{balakrishnan_constrained_2019} enhance E2E MRs with more detail on the target linguistic structure, which they then use to constrain decoding.

Since its initial release in \citep{Novikova:Sigdial2017}, the E2E dataset has motivated several authors to explore generating more diverse outputs, mostly with additional supervision signals:
The system of \citet{wiseman_learning_2018} learns latent templates (sequences of phrases/slots) while learning to generate, thus allowing more controllability and arguably more diversity of the outputs -- the templates serve as an additional, fine-grained way of specifying the desired shape of the generator output.
\citet{reed_can_2018} explore using the presence of prespecified contrast markers (e.g.\ \emph{but}, \emph{although}) as additional supervision, while \citet{juraska_characterizing_2018} investigate other stylistic markers and use them to generate sentences of specified type. \citet{oraby_controlling_2018} and \citet{oraby_neural_2018} attempt to generate outputs showing different personality traits (represented by the Big Five model) using additional synthetic training data with personality annotation. In an extension of their E2E competing system \citep{deriu_end--end_2018}, \citet{deriu_syntactic_2018} add specific syntactic features to the input MRs to control not only the first word of the output, but also first words of all sentences in multi-sentence outputs and specific phrasing for expressing each slot in the MR. \citet{jagfeld_sequence--sequence_2018} do not add more supervision but compare the diversity produced by word-level and character-level seq2seq models on E2E data, showing better performance of the latter.

Using an in-house restaurant dataset, \citet{nayak_plan_2017} explore using a basic sentence plan specification (slot ordering and sentence grouping) as an additional training signal to increase output diversity. Working in the transport information domain, \citet{dusek_context-aware_2016} and \citet{mangrulkar_context-aware_2018} condition their generators on preceding dialogue context as well as the input MR to obtain greater diversity.

\section{Evaluation Setup}\label{sec:eval}

We evaluated the systems submitted to the E2E challenge using a range of automatic metrics, which we describe in Section~\ref{sec:automatic-metrics}. This includes a novel application of textual measures\footnote{These measures were previously applied by \citet{Perez-Beltrachini17} and this work (see Section~\ref{sec:dataset}) to describe datasets, but not for evaluation of NLG outputs.} and a novel usage of standard word-overlap metrics to assess similarity among individual systems.
Automatic metrics are popular in NLG \citep{gkatzia_snapshot_2015} because they are cheaper and faster to run than human evaluation. However, sole use of automatic metrics is only sensible if they are known to be sufficiently correlated with human preferences.
Recent studies~\citep{Novikova:EMNLP2017,reiter_structured_2018} have demonstrated that this is very often not the case and that automatic metrics only weakly reflect human judgements on system outputs as generated by data-driven NLG.
Therefore, we also performed a large-scale crowdsourced human evaluation, as detailed in Section~\ref{sec:human-evaluation}.
For the human evaluation of the 20 primary systems, we address the problem of how to efficiently compare a large number of systems, by:
\begin{enumerate}
\item Extending our previous work \citep{novikova:2018} on rank-based Magnitude Estimation (RankME) and verifying the method at scale;\footnote{The original study \citep{novikova:2018} was limited to comparing 3 similar systems on 100 utterances.}
\item Introducing the data-efficient TrueSkill algorithm \citep{herbrich2007trueskill,sakaguchi2014efficient} to NLG. This allows us to compute an overall ranking by directly comparing the systems, rather than individually assessing them at higher cost, as done by previous NLG challenges \citep{belz:2014}.
\end{enumerate}

\subsection{Automatic Metrics}\label{sec:automatic-metrics}

We apply two types of automatic metrics: One set assessing the similarity between generated system outputs and natural language references in the corpus using word-overlap-based measures, and another set assessing the complexity and diversity of system outputs using a variety of textual measures.

\subsubsection*{Word-overlap metrics}

For the first set, we selected a range of
metrics measuring word-overlap between system output and references, including 
BLEU and NIST, which are used as standard in machine translation evaluation \citep{bojar_results_2017} and very common in NLG, and several others which were applied in the COCO caption generation challenge \citep{chen_microsoft_2015} as well as other NLG experiments \cite[e.g.][]{lebret_neural_2016,gardent_webnlg_2017,sharma_natural_2016}:
\begin{description}

\item[\bleu \rm\citep{papineni2002bleu}] is the harmonic mean of $n$-gram precisions of the system output with respect to human-authored reference sentences, with $n\in\{1,\dots,4\}$, lowered by a brevity penalty if the output is shorter than references. The $n$-gram precisions are proportions of $n$-grams in the system output that can be matched in any of the reference sentences. Repeated $n$-gram matches are clipped to the maximum number of times the $n$-gram occurs in any single reference.

\item[\nist \rm\citep{nist}] is a version of \bleu with higher weighting for less frequent (i.e., more informative) $n$-grams and a different length penalty. It uses $n\in\{1,\dots,5\}$.

\item[\meteor \rm\citep{lavie_meteor:_2007}] measures both precision and recall of unigrams by aligning the system output with the individual human references. In addition to exact word matches, it uses fuzzy matching based on stemming and WordNet synonyms. It computes matches against multiple references separately and uses the best-matching one.

\item[\rouge \rm\citep{lin2004rouge}] is based on longest common subsequences (LCS) between the system output and the human references, where a common subsequence requires the same words in the same order but allows additional, non-covered words in the middle of either sequence. The final \rouge score is an F-measure based on maximum precision and maximum recall achieved over any of the human references, where precision and recall are computed as length of the LCS divided by the length of the system output and the reference, respectively.

\item[\cider \rm\citep{cider}] was primarily designed for generated image captions, but is also applicable for NLG in general. \cider is computed as the average cosine similarity between the system output and the reference sentences on the level of $n$-grams, $n\in \{1,\dots,4\}$. The importance of the individual $n$-grams is given by the Term Frequency Inverse Document Frequency (TF-IDF) measure, which weighs an $n$-gram's frequency in a particular instance against its overall frequency in the whole dataset.

\end{description}
We provided scripts to the challenge participants to run all of these metrics in a simple, easy-to-use way. The scripts are freely available at the following URL:\footnote{The scripts are partially based on COCO caption generation challenge evaluation scripts (\url{https://github.com/tylin/coco-caption}).}
\begin{center}
\url{https://github.com/tuetschek/e2e-metrics}
\end{center}

In addition to evaluating all NLG systems individually against human-authored reference texts (see Section~\ref{sec:results-automatic}), we also apply the same metrics as measures of output similarity among the systems, comparing each system's outputs with all other systems' outputs in place of references (see Section~\ref{sec:systems-similarity}).

\subsubsection*{Textual metrics}

For the second set of scores, which is intended to measure complexity and diversity in the system outputs, we use the same automatic textual metrics which we
 used to evaluate the E2E NLG dataset itself (see Section~\ref{sec:lexical-richness} and~\ref{sec:syntactic-variation}), i.e.\ dimensions of lexical richness, such as lexical sophistication (LS2) and mean segmental token-to-type ratio (MSTTR), and metrics of syntactic complexity, such as levels of the revised D-level Scale.\footnote{The script used to install all third-party tools required and run the comparison is available at \url{https://github.com/tuetschek/e2e-stats}.}
This allows us to both evaluate the diversity and complexity of system outputs and to establish whether the text characteristics are similar to the training and test sets.
To focus specifically on the style produced by the individual systems, we delexicalized restaurant names in the system outputs before computing textual metrics scores, since restaurant names could skew some of these metrics as they are mostly composed of infrequent nouns (cf.\ Section~\ref{sec:lexical-richness}).

\subsection{Human Evaluation}\label{sec:human-evaluation}

The human evaluation was conducted on the 20 primary systems and the baseline using 
Rank-based Magnitude Estimation (RankME)~\citep{novikova:2018}.
In an ordinary (i.e.\ not rank-based) ME task~\citep{Bard:1996}, subjects provide a relative rating of an experimental sentence to a reference sentence, which is associated with a pre-set/fixed number. If the target sentence appears twice as good as the reference sentence, for instance, subjects are to multiply the reference score by two; if it appears half as good, they should divide it in half, etc. Rank-based ME extends this idea by asking subjects to provide a relative ranking of several target sentences, i.e. not only to the reference sentence, but also to each other.

Rank-based ME was selected for several reasons.
First, its use proved to significantly increase the consistency of human ratings, compared to other data collection methods~\citep{novikova:2018}.
Second, it implies the use of continuous scales, i.e.\ rating scales without numerical labels and without given end points. Recent studies show that continuous scales allow subjects to give more nuanced judgements \citep{belz2011discrete,graham_continuous_2013,bojar2017findings}.
 Third, it explores relative ranking of different systems instead of directly assessing quality of each specific system, which makes it more reliable in the environment of a challenge.

The evaluation was conducted using crowdsourcing based on the CrowdFlower/\hspace{0mm}FigureEight platform.  Crowd workers were presented with five randomly selected outputs of different systems corresponding to a single MR, and were asked to evaluate and rank these systems from the best to the worst, ties permitted, using the RankME method.

The final evaluation results were produced using the TrueSkill algorithm ~\citep{herbrich2007trueskill, sakaguchi2014efficient}. TrueSkill produces system rankings by gradually updating a Bayesian estimate of each system's capability according to the ``surprisal'' of pairwise comparisons of individual system outputs. This way, fewer direct comparisons between systems are needed to establish their overall ranking. In \citep{novikova:2018}, we were able to show that TrueSkill is able to to reduce the amount of collected human evaluation data without compromising the final ranking results.

Since the performance of some systems may be very similar and a total ordering would not reflect this, we adopt the practice used in machine translation of presenting a partial ordering into significance clusters established by bootstrap resampling \citep{bojar_findings_2013,bojar_findings_2014,sakaguchi2014efficient}. The TrueSkill algorithm is run 200 times, producing slightly different rankings each time as pairs of system outputs for comparison are randomly sampled. This way we can determine the range of ranks where each system is placed 95\% of the time or more often. Clusters are then formed of systems whose rank ranges overlap.

Traditionally, human evaluation aims to assess the naturalness (fluency, readability) and informativeness (relevance, correctness, adequacy) of an automatically generated output \citep{gatt2017survey}.
Naturalness targets the linguistic quality of the NLG system output;
informativeness targets relevance or correctness of the output relative to the input MR, showing how well the system reflects the MR content. Recent
research often adds a general, overall quality criterion \citep{wen:emnlp2015,wen_stochastic_2015,manishina_automatic_2016,novikova:INLG2016,Novikova:EMNLP2017}, or even uses only that \citep{SharmaHSSB16}.

We decided against explicitly evaluating informativeness 
 since our training instances do not always verbalise all MR attributes (cf. Section \ref{sec:content-selection}).
We therefore only collected separate ranks for \textit{quality} and \textit{naturalness}.
\begin{description}
\item[Quality:] 
When collecting quality ratings, system outputs were presented to crowd workers together with the corresponding meaning representation, which implies that correctness of the NL utterance relative to the MR should also influence this ranking.
The crowd workers were asked: ``{\em How do you judge the overall quality of the utterance in terms of its grammatical correctness, fluency, adequacy and other important factors?}"

\item[Naturalness:] 
When collecting naturalness ratings, system outputs were presented to crowd workers without the corresponding meaning representation. The crowd workers were asked: ``{\em Could the utterance have been produced by a native speaker?}"
\end{description}

Ratings of quality and naturalness were collected separately, i.e.\ in two individual crowdsourcing tasks. Furthermore, when crowd workers were asked to assess naturalness, the MR was not shown to them since it was not necessary for the task. This setup allows to minimise the correlation between the ratings of naturalness and quality~\citep{novikova:2018,callison-burch_meta-_2007}.

\section{Results}\label{sec:results}

In this section, we report on the results of the evaluation of all E2E NLG Challenge primary systems, following the evaluation procedures described in Section~\ref{sec:eval}. We first show the results using automatic metrics: word-overlap-based (Section~\ref{sec:results-automatic}) and textual metrics (Section~\ref{sec:results-text-metrics}), as well as automatically computed output similarity between systems (Section~\ref{sec:systems-similarity}).
We then summarise 
 the human evaluation results (Section~\ref{sec:results-human}), comment on the semantic accuracy of system outputs (Section~\ref{sec:coverage}) and declare the overall winning system (Section~\ref{sec:winner}).
Finally, we provide a list of ``lessons learnt'' in Section~\ref{sec:lessons} -- observations that we hope will be useful for future NLG system development.

\subsection{Word-overlap Metrics}
\label{sec:results-automatic}

\begin{table}[tbp]
\setlength{\extrarowheight}{3pt}
\begin{center}
\small
\begin{tabular}{llccccccc}\hline
\textbf{System} & \bf BLEU & \bf NIST & \bf \hspace{-2mm}METEOR\hspace{-1mm} & \bf \hspace{-2mm}ROUGE-L\hspace{-3mm} & \bf CIDEr & \bf norm.~avg. \\ \hline
\Ctgen  & 0.6593  & 8.6094  & 0.4483  & 0.6850  & 2.2338 & 0.5754 \\\hdashline[0.5pt/2pt]
\Cslug  & \bf 0.6619  & \bf 8.6130  & 0.4454  & 0.6772  & \bf 2.2615 & 0.5744 \\
\Ctntnlgi  & 0.6561  & 8.5105  & \bf 0.4517  & 0.6839  & 2.2183 & 0.5729 \\
\Cnle & 0.6534  & 8.5300  & 0.4435  & 0.6829  & 2.1539 & 0.5696 \\
\Ctntnlgii  & 0.6502  & 8.5211  & 0.4396  & \bf 0.6853  & 2.1670 & 0.5688 \\
\Charv  & 0.6496  & 8.5268  & 0.4386  & \bf 0.6872  & 2.0850 & 0.5673 \\
\Czhang  & 0.6545  & 8.1840  & 0.4392  &\bf 0.7083  & 2.1012 & 0.5661 \\
\Cgong  & 0.6422  & 8.3453  & 0.4469  & 0.6645  & \bf 2.2721 & 0.5631 \\
\Cthomsoni & 0.6336  & 8.1848  & 0.4322  & 0.6828  & 2.1425 & 0.5563 \\
\Csheffi & 0.6015  & 8.3075  & 0.4405  & 0.6778  & 2.1775 & 0.5537 \\
\Cdangnt  & 0.5990  & 7.9277  & 0.4346  & 0.6634  & 2.0783 & 0.5395 \\
\Cslugalt~\it (late submission)  & 0.6035  & 8.3954  & 0.4369  & 0.5991  & 2.1019 & 0.5378 \\
\Czhawii  & 0.6004  & 8.1394  & 0.4388  & 0.6119  & 1.9188 & 0.5314 \\
\Ctuda  & 0.5657  & 7.4544  & \bf 0.4529  & 0.6614  & 1.8206 & 0.5215 \\
\Czhawi  & 0.5864  & 8.0212  & 0.4322  & 0.5998  & 1.8173 & 0.5205 \\
\Cadapt  & 0.5092  & 7.1954  & 0.4025  & 0.5872  & 1.5039 & 0.4738 \\
\Cchen & 0.5859  & 5.4383  & 0.3836  & 0.6714  & 1.5790 & 0.4685 \\
\Cforgeiii  & 0.4599  & 7.1092  & 0.3858  & 0.5611  & 1.5586 & 0.4547 \\
\Csheffii  & 0.5436  & 5.7462  & 0.3561  & 0.6152  & 1.4130 & 0.4462 \\
\Cthomsonii  & 0.4202  & 6.7686  & 0.3968  & 0.5481  & 1.4389 & 0.4372 \\
\Cforgei  & 0.4207  & 6.5139  & 0.3685  & 0.5437  & 1.3106 & 0.4231 \\\hline
\end{tabular}
\end{center}
\caption{Word-overlap metrics scores (see Section~\ref{sec:automatic-metrics}) for all primary systems, plus the average of all metrics' values normalised into the 0-1 range. The list is sorted by the normalised average; any values higher than the corresponding baseline are marked in bold. 
System architectures are coded with colours and symbols: \textcolor{seqtoseq}{\symbseq seq2seq}, \textcolor{datadriven}{\symbdd other data-driven}, \textcolor{rules}{\symbrule rule-based}, \textcolor{templates}{\symbtempl template-based}.}
\label{tab:primary-systems-wbms}
\end{table}

Table~\ref{tab:primary-systems-wbms} summarises the system scores for word-overlap metrics (cf.~Section~\ref{sec:automatic-metrics}).
It is apparent that the \tgen baseline is very strong in terms of word-overlap metrics: No primary system is able to beat it in terms of all metrics, or in terms of the normalised metrics' mean -- only \slug comes very close. Several other systems manage to beat \tgen in one of the metrics but not in others. Note, however, that many secondary system submissions perform better than the primary ones (and the baseline) with respect to word-overlap metrics (see Table~\ref{tab:all-systems-wbms} in the Appendix).

Overall, seq2seq-based systems show the best word-based metric values, followed by \sheffi, a data-driven system based on imitation learning. As expected, attempts to increase output diversity by \zhawi, \zhawii, \slugalt and \adapt result in lowered scores by word-overlap-based metrics. 
Template-based and rule-based systems mostly score at the bottom of the list.  The lowest-scoring systems in terms of word-overlap metrics are the ones of \chen and \sheffii, which tend to produce much shorter outputs than other systems (cf.~Section~\ref{sec:results-text-metrics}). This most likely 
resulted in severe brevity penalty.

Finally, it must be noted that the results using automatic metrics are quite different from results obtained in human evaluation (see Section~\ref{sec:results-human}), which confirms previous findings \citep{Novikova:EMNLP2017,reiter_structured_2018}.

\subsection{Textual Metrics}\label{sec:results-text-metrics}

\begin{table}[tb]
\begin{center}
\setlength{\extrarowheight}{3pt}
\scriptsize
\begin{tabular}{l@{\hskip 0.1cm}r@{\hskip 0.7cm}l@{\hskip 0.1cm}r@{\hskip 0.7cm}l@{\hskip 0.1cm}r@{\hskip 0.7cm}l@{\hskip 0.1cm}r@{\hskip 0.7cm}l@{\hskip 0.1cm}r@{\hskip 0.7cm}l@{\hskip 0.1cm}r}
\hline
\multicolumn{2}{c@{\hskip 0.7cm}}{\bf \% Level0-2 } & \multicolumn{2}{c@{\hskip 0.7cm}}{\bf \% Level6-7 } & \multicolumn{2}{c@{\hskip 0.7cm}}{\bf LS2} & \multicolumn{2}{c@{\hskip 0.7cm}}{\bf MSTTR-50} & \multicolumn{2}{c}{\bf Avg.\ length}\\
\hline
\Cgong    & 82.68    & \Csheffi    & 41.27    & \it test set all    & 0.43    & \it test set rand    & 0.62    & \Ctuda    & 31.02 \\
\Ctntnlgii    & 79.64    & \Cforgei    & 33.66    & \it test set rand    & 0.36    & \Cthomsonii    & 0.62    & \Cthomsonii    & 27.48 \\
\Cslug    & 78.08    & \Cslugalt    & 30.49    & \Cadapt    & 0.33    & \Cadapt    & 0.61    & \Cforgei    & 26.88 \\
\Ctntnlgi    & 72.18    & \Czhawi    & 26.00    & \Cforgei    & 0.30    & \Cforgei    & 0.59    & \Czhawii    & 26.58 \\
\Czhang    & 70.83    & \Cthomsonii    & 21.07    & \Cthomsonii    & 0.29    & \Czhawi    & 0.58    & \Ctntnlgi    & 26.37 \\
\Cdangnt    & 66.95    & \Czhawii    & 19.03    & \Charv    & 0.27    & \it test set all    & 0.58    & \Czhawi    & 26.16 \\
\Ctgen    & 65.12    & \Cforgeiii    & 18.51    & \Ctntnlgi    & 0.26    & \Czhawii    & 0.57    & \Ctntnlgii    & 25.49 \\
\Charv    & 64.63    & \it test set rand    & 17.46    & \Cchen    & 0.25    & \Cforgeiii    & 0.56    & \Cgong    & 25.41 \\
\Cthomsoni    & 64.28    & \Cgong    & 16.90    & \Cnle    & 0.25    & \Ctuda    & 0.55    & \Cdangnt    & 24.85 \\
\Cforgeiii    & 62.62    & \it test set all    & 16.48    & \Csheffii    & 0.25    & \Cdangnt    & 0.54    & \Cadapt    & 24.47 \\
\Cadapt    & 62.48    & \Cslug    & 11.39    & \Csheffi    & 0.24    & \Cslugalt    & 0.54    & \Cslugalt    & 24.47 \\
\Cforgei    & 61.13    & \Cnle    & 11.12    & \Ctntnlgii    & 0.23    & \Cslug    & 0.52    & \it test set rand    & 24.39 \\
\Czhawi    & 58.91    & \Ctuda    & 10.48    & \Ctgen    & 0.22    & \Ctntnlgi    & 0.52    & \Ctgen    & 24.04 \\
\Cnle    & 58.24    & \Cadapt    & 10.28    & \Cdangnt    & 0.21    & \Csheffi    & 0.52    & \it test set all    & 23.96 \\
\it test set rand    & 58.16    & \Ctntnlgi    & \phantom{0}9.55    & \Ctuda    & 0.21    & \Cnle    & 0.52    & \Cslug    & 23.76 \\
\it test set all    & 57.97    & \Ctgen    & \phantom{0}9.02    & \Cthomsoni    & 0.20    & \Ctgen    & 0.52    & \Cforgeiii    & 23.49 \\
\Ctuda    & 57.66    & \Cdangnt    & \phantom{0}8.91    & \Czhang    & 0.20    & \Ctntnlgii    & 0.51    & \Cnle    & 23.40 \\
\Cthomsonii    & 57.36    & \Cthomsoni    & \phantom{0}8.13    & \Cslug    & 0.20    & \Charv    & 0.51    & \Charv    & 23.22 \\
\Cchen    & 54.35    & \Charv    & \phantom{0}8.12    & \Cgong    & 0.20    & \Cthomsoni    & 0.50    & \Csheffi    & 22.75 \\
\Csheffii    & 52.98    & \Czhang    & \phantom{0}5.27    & \Cforgeiii    & 0.20    & \Cgong    & 0.50    & \Cthomsoni    & 22.43 \\
\Czhawii    & 52.63    & \Ctntnlgii    & \phantom{0}5.22    & \Cslugalt    & 0.19    & \Czhang    & 0.47    & \Czhang    & 20.71 \\
\Cslugalt    & 35.12    & \Cchen    & \phantom{0}4.40    & \Czhawii    & 0.17    & \Cchen    & 0.43    & \Csheffii    & 17.18 \\
\Csheffi    & 26.19    & \Csheffii    & \phantom{0}2.08    & \Czhawi    & 0.17    & \Csheffii    & 0.43    & \Cchen    & 16.32 \\
\hline
\end{tabular}
\end{center}
\caption{Systems sorted according to selected textual metrics (percentage of simple and complex sentences, lexical sophistication LS2, MSTTR-50, average output length in tokens). For comparison, the table also includes the same values for the whole test set (\emph{test set all}) and for a randomly selected subset of the test set, with one reference text per MR (\emph{test set rand}).
System architectures are coded with colours and symbols: \textcolor{seqtoseq}{\symbseq seq2seq}, \textcolor{datadriven}{\symbdd other data-driven}, \textcolor{rules}{\symbrule rule-based}, \textcolor{templates}{\symbtempl template-based}.
}\label{tab:systems-textual}
\end{table}

\begin{table}[tb]
\begin{center}
\setlength{\extrarowheight}{3pt}
\scriptsize
\begin{tabular}{l@{\hskip 0.1cm}r@{\hskip 0.7cm}l@{\hskip 0.1cm}r@{\hskip 0.7cm}l@{\hskip 0.1cm}r@{\hskip 0.7cm}l@{\hskip 0.1cm}r@{\hskip 0.7cm}l@{\hskip 0.1cm}r@{\hskip 0.7cm}l@{\hskip 0.1cm}r}
\hline
\multicolumn{2}{c@{\hskip 0.7cm}}{\bf Distinct tokens} & \multicolumn{2}{c@{\hskip 0.7cm}}{\hspace{-0.3cm}\bf Distinct trigrams} & \multicolumn{2}{c@{\hskip 0.7cm}}{\hspace{-0.3cm}\bf \% Unique trigrams} & \multicolumn{2}{c@{\hskip 0.7cm}}{\hspace{-0.2cm}\bf Entropy tokens} & \multicolumn{2}{c}{\hspace{-0.5cm}\bf Cond.\ entropy bigrams}\\
\hline
\it test set all    & 1079    & \it test set all    & 16797    & \it test set rand    & 69.13    & \it test set all    & 6.40    & \it test set all    & 2.92 \\
\it test set rand    & 542    & \it test set rand    & 5166    & \Cadapt    & 66.61    & \it test set rand    & 6.37    & \it test set rand    & 2.70 \\
\Cadapt    & 455    & \Cthomsonii    & 4687    & \Cthomsonii    & 60.44    & \Cthomsonii    & 6.24    & \Cthomsonii    & 2.60 \\
\Cthomsonii    & 399    & \Cadapt    & 3567    & \it test set all    & 44.66    & \Cadapt    & 6.18    & \Cadapt    & 2.09 \\
\Czhawi    & 136    & \Czhawi    & 969    & \Czhawi    & 24.97    & \Cforgeiii    & 5.74    & \Cforgeiii    & 1.66 \\
\Cforgeiii    & 124    & \Cforgeiii    & 896    & \Charv    & 21.88    & \Czhawi    & 5.71    & \Cslugalt    & 1.55 \\
\Czhawii    & 102    & \Cslugalt    & 855    & \Ctntnlgi    & 21.34    & \Czhawii    & 5.65    & \Charv    & 1.45 \\
\Charv    & 93    & \Charv    & 777    & \Cnle    & 18.75    & \Cslugalt    & 5.57    & \Czhawi    & 1.44 \\
\Ctntnlgi    & 89    & \Czhawii    & 716    & \Czhawii    & 18.72    & \Cforgei    & 5.55    & \Ctntnlgii    & 1.39 \\
\Cforgei    & 88    & \Ctntnlgi    & 703    & \Cslugalt    & 18.13    & \Charv    & 5.50    & \Cnle    & 1.37 \\
\Cslugalt    & 88    & \Ctntnlgii    & 634    & \Cchen    & 17.92    & \Csheffi    & 5.43    & \Ctntnlgi    & 1.37 \\
\Ctntnlgii    & 86    & \Cnle    & 608    & \Czhang    & 17.81    & \Cnle    & 5.43    & \Csheffi    & 1.33 \\
\Ctgen    & 83    & \Ctgen    & 597    & \Csheffi    & 16.44    & \Ctgen    & 5.41    & \Ctgen    & 1.32 \\
\Cnle    & 81    & \Csheffi    & 578    & \Cslug    & 15.58    & \Ctntnlgi    & 5.37    & \Czhawii    & 1.32 \\
\Czhang    & 76    & \Cforgei    & 549    & \Cforgeiii    & 13.50    & \Cslug    & 5.35    & \Cthomsoni    & 1.30 \\
\Cthomsoni    & 75    & \Czhang    & 511    & \Ctgen    & 13.23    & \Ctntnlgii    & 5.34    & \Cforgei    & 1.29 \\
\Cslug    & 74    & \Cslug    & 507    & \Ctntnlgii    & 12.93    & \Cdangnt    & 5.29    & \Czhang    & 1.26 \\
\Cchen    & 73    & \Cchen    & 480    & \Cforgei    & 12.39    & \Ctuda    & 5.25    & \Cchen    & 1.17 \\
\Csheffi    & 72    & \Cthomsoni    & 464    & \Cthomsoni    & 10.78    & \Cthomsoni    & 5.24    & \Cslug    & 1.13 \\
\Cdangnt    & 61    & \Cdangnt    & 301    & \Cgong    & 7.30    & \Czhang    & 5.21    & \Csheffii    & 1.10 \\
\Csheffii    & 59    & \Csheffii    & 262    & \Csheffii    & 4.96    & \Cgong    & 5.19    & \Cdangnt    & 1.06 \\
\Cgong    & 58    & \Cgong    & 233    & \Cdangnt    & 0.00    & \Cchen    & 5.09    & \Cgong    & 0.91 \\
\Ctuda    & 57    & \Ctuda    & 143    & \Ctuda    & 0.00    & \Csheffii    & 4.76    & \Ctuda    & 0.71 \\
\hline
\end{tabular}
\end{center}
\caption{Systems sorted according to selected textual diversity metrics (number of distinct tokens, number fo distinct trigrams, proportion of unique trigrams, Shannon entropy over tokens (unigrams), bigram next-word conditional entropy). For comparison, the table also includes the same values for the whole test set (\emph{test set all}) and for a randomly selected subset of the test set, with one reference text per MR (\emph{test set rand}).
System architectures are coded with colours and symbols: \textcolor{seqtoseq}{\symbseq seq2seq}, \textcolor{datadriven}{\symbdd other data-driven}, \textcolor{rules}{\symbrule rule-based}, \textcolor{templates}{\symbtempl template-based}.
}\label{tab:systems-entropy}
\end{table}

Table~\ref{tab:systems-textual} summarises results from a range of textual metrics which aim to assess the complexity and diversity of primary system outputs (cf.~Section~\ref{sec:automatic-metrics}).
In addition, we include a comparison to the human references in the test set in order to assess whether systems are able to replicate characteristics of human-produced data.\footnote{Note that textual metrics have been computed with restaurant names delexicalised (cf.~Section~\ref{sec:automatic-metrics}).}
The results in Table~\ref{tab:systems-textual} show the following:
\begin{itemize}
\item Seq2seq-based system outputs are less syntactically complex on average than outputs of other systems (they produce more D-level 0-2 sentences and less D-level 6-7 sentences than other architectures).

\item The systems seem to show a relatively high variance in syntactic complexity levels,
especially with respect to the higher levels; few systems match the distribution of the
training and test data.
The differences in D-level distributions in the outputs are mostly statistically significant (see Figure~\ref{tab:diff_dlevels} in the Appendix).
The only system producing a D-level distribution \emph{not} significantly different from a random test set reference is \forgeiii, which is based on template mining from training data.

If we use Bhattacharyya distance to compare the D-level distributions (cf.~Figure~\ref{tab:dist_dlevels} in the Appendix), the greatest distances appear in both extremes. \sheffi, \forgei and \slugalt produce higher-level sentences more frequently and thus show among the most distant from other systems. The \gong system mostly produces level 0-2 sentences, and therefore it appears very distant from other systems as well as the most distant system from human references.

\item None of the systems reaches the lexical sophistication of the human-authored test set references. The diversity-attempting seq2seq-based \adapt system comes very close, followed by the grammar-based \forgei and the \thomsonii system, which is based on template mining from data. Data-driven systems aiming at higher lexical diversity seem to achieve higher sophistication as well; note the lower performance of \slugalt, which aims more at syntactic diversity than lexical. For rule-based systems, lexical sophistication is a direct result of the system authors' decisions.

\item In terms of MSTTR, highest scores are achieved by template or rule-based systems and by data-driven systems that explicitly aim at greater output diversity (\zhawi, \zhawii, \adapt, \slugalt). Note that MSTTR is typically higher in systems that tend to produce longer outputs, which includes most rule- and template-based systems. 
We assume that this is due to MSTTR's fixed 50-token window used to segment utterances.

\item Most systems produce outputs similar in length to the test set human references. Outputs of rule- and template-based systems tend to be more verbose than those of data-driven systems. The outputs of \zhang, \sheffii and \chen are much shorter on average than texts in the dataset, which suggests that these systems might not verbalise all the information contained in the MR (cf.~Section~\ref{sec:coverage}).
\end{itemize}

Same as for the datasets statistics in Section~\ref{sec:lexical-richness}, we also computed additional textual measures to assess the diversity/repetitiveness of the generated outputs: number of distinct n-grams, Shannon entropy, and conditional next-word entropy; a selection of these metrics is shown in Table~\ref{tab:systems-entropy}.\footnote{We used system outputs with delexicalised restaurant names for the evaluation, but the lexicalised outputs show the same trends. The values for n-gram lengths not displayed in Table~\ref{tab:systems-entropy} also show very similar trends.} We compare the outputs against the whole test set (multiple references) and a randomly selected single reference per MR from the test set. The results show the following:

\begin{itemize}
\item
None of the systems is able to produce as much diversity as is contained in a randomly selected human reference -- even the most diverse systems lag behind. \adapt comes close in vocabulary size, \thomsonii is the closest system in terms of entropy and next-word conditional entropy.

\item
In terms of vocabulary, there is a huge gap between the most diverse \adapt and \thomsonii systems, and any other system (e.g., the 3rd-ranking \zhawi has 3$\times$ smaller vocabulary than \thomsonii, and 2.4$\times$ smaller ratio of unique trigrams).

\thomsonii demonstrates that mining templates from the training data can lead to very diverse outputs. \forgeiii, which uses the same method, also ranks relatively high on vocabulary size and entropy. The diversity produced by \adapt's seq2seq model indicates that the prepocessing step enriching the MRs works effectively (cf.\ Section~\ref{sec:data-augmentation}).

\item
All diversity-attempting data-driven systems (\adapt, \zhawi, \zhawii, \harv, \tntnlgi, \tntnlgii, \slugalt) indeed rank better than most systems not incorporating diversity measures, with \tntnlgi and \tntnlgii showing lower gains than the rest of the group. However, template-mining-based systems (\thomsonii, \forgeiii) produce outputs of similar or higher diversity with no concentrated effort.

\item
Outputs of seq2seq-based systems which do not explicitly model diversity (e.g. \gong, \sheffi, \thomsoni, \slug, \chen) indeed show lower diversity  scores.
The rule-based \dangnt system also ranks very low on diversity, and the \tuda system with handcrafted templates is the least diverse of all.

\end{itemize}

In summary,
few systems are able to approach the complexity and diversity shown in human-authored data. Seq2seq-based systems tend to favor simpler sentences than hand-engineered systems unless diversity control is in place. Vanilla seq2seq and handcrafted templates produce the least diverse outputs; highest diversity is achieved by template mining or explicit diversity control mechanisms.

\subsection{System Output Similarity}\label{sec:systems-similarity}

\begin{figure}[tbp]
\centering
\scriptsize
\setlength{\extrarowheight}{1pt}
\includegraphics[height=8.5cm]{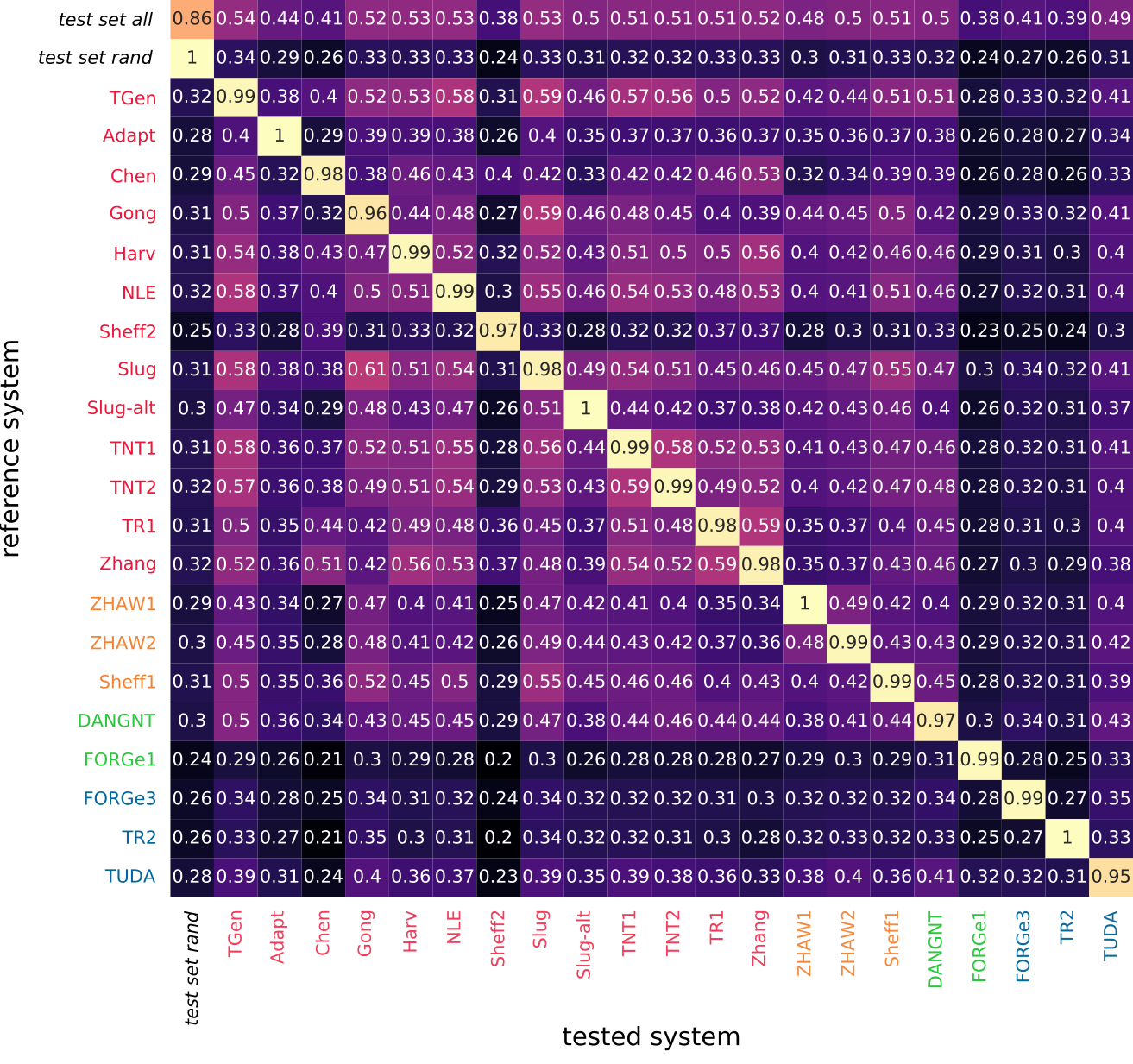}
\hfill
\begin{tabular}[b]{lr}\hline
\bf System & \bf Mean \\\hline
\Ctgen & 0.48 \\
\Cslug & 0.47 \\
\Ctntnlgi & 0.46 \\
\Cnle & 0.46 \\
\Ctntnlgii & 0.46 \\
\Charv & 0.46 \\
\Czhang & 0.45 \\
\Csheffi & 0.44 \\
\Cthomsoni & 0.44 \\
\Cgong & 0.44 \\
\Cdangnt & 0.42 \\
\Cslugalt & 0.42 \\
\Czhawii & 0.42 \\
\Czhawi & 0.40 \\
\Cchen & 0.40 \\
\Ctuda & 0.37 \\
\Cadapt & 0.37 \\
\Cforgeiii & 0.34 \\
\Csheffii & 0.34 \\
\it test set rand & 0.34 \\
\Cthomsonii & 0.33 \\
\Cforgei & 0.31 \\\hline
\\ 
\\
\\
\\
\end{tabular}
\caption{Similarity of the systems' outputs as measured by automatic metrics (mean of normalised \bleu, \nist, \meteor, \rouge and \cider where one system output is used as reference). Systems are sorted by their architecture. For comparison, we also include metrics values against the full test set with multiple references (\emph{test set all}) and against a single-reference randomly sampled subset of the test set (\emph{test set rand}). The table on the right shows mean values of similarity of each system against all other systems (average over columns on the left, excluding the 1st line).
System architectures are coded with colours and symbols: \textcolor{seqtoseq}{\symbseq seq2seq}, \textcolor{datadriven}{\symbdd other data-driven}, \textcolor{rules}{\symbrule rule-based}, \textcolor{templates}{\symbtempl template-based}.
}
\label{fig:metrics-heatmap}
\end{figure}

In order to assess the similarity of outputs produced by the individual systems, we reused the word-overlap-based metrics applied in the challenge (see Section~\ref{sec:automatic-metrics}).
We created all possible pairs of systems and computed word-overlap metrics between each of their outputs for every instance in the test set.
Same as for textual metrics, restaurant names were delexicalised in the system outputs.
\footnote{Results with fully lexicalised outputs are very similar, the differences are just slightly less profound.}

This process resulted in a table for each of the metrics (see Figure~\ref{fig:metrics-heatmap-all} in the Appendix), with reference systems in rows and tested systems in columns.
All five metrics showed a very similar pattern. Figure~\ref{fig:metrics-heatmap} therefore summarises the results by taking the average of all normalised metrics (cf.~Table~\ref{tab:primary-systems-wbms}).
For comparison, we also measure similarity of
 system outputs against the reference texts in the test set, as well as a subset of the test set with a single, randomly sampled reference text per MR.

We can see from Figure~\ref{fig:metrics-heatmap} that all the seq2seq-based system outputs are in general most similar to each other; other data-driven systems also show higher similarity amongst each other.
The exception to this rule in case of the \chen and \sheffii systems can be explained by the brevity of their outputs (cf.~Sections~\ref{sec:results-automatic} and~\ref{sec:results-text-metrics}). Systems that aim at output diversity (\zhawi, \zhawii, \slugalt and mainly \adapt) also exhibit lowered similarity of their outputs to those of other systems, which might indicate that their outputs are indeed more original. The outputs of rule-based and template-based systems are markedly less similar to other outputs than that of the data-driven systems.

We can also see that most system outputs, especially those of data-driven methods, are much more similar to each other than they are to a single randomly selected human-authored reference text from the test set. This is to be expected since data-driven methods tend to select more frequent phrasing.
Some of the system outputs even show a higher similarity to each other than to the closest matching human references from the test set. This is mainly the case for systems with very similar architectures, which often arrive at identical results (e.g.\ \tgen, \tntnlgi and \tntnlgii).

\subsection{Results of Human Evaluation}

\label{sec:results-human}

\begin{table}[tbp]
\begin{center}
\footnotesize
\setlength{\extrarowheight}{2pt}
\begin{tabular}{ccr@{--}ll}
\multicolumn{5}{c}{\normalsize\bf Quality\Bstrut}\\\hline
\bf \#  & \bf TrueSkill  & \multicolumn{2}{c}{\bf Rank}  & \bf System \\\hline
1  & \phantom{-}0.300  & 1 & 1  & \Cslug \\\hdashline[0.5pt/2pt]
\multirow{13}{*}{2}  & \phantom{-}0.228  & 2 & 4  & \Ctuda \\
& \phantom{-}0.213  & 2 & 5  & \Cgong \\
& \phantom{-}0.184  & 3 & 5  & \Cdangnt \\
& \phantom{-}0.184  & 3 & 6  & \Ctgen \\
& \phantom{-}0.136  & 5 & 7  & \Cslugalt~\emph{(late)} \\
& \phantom{-}0.117  & 6 & 8  & \Czhawii \\
& \phantom{-}0.084  & 7 & 10  & \Ctntnlgi \\
& \phantom{-}0.065  & 8 & 10  & \Ctntnlgii \\
& \phantom{-}0.048  & 8 & 12  & \Cnle \\
& \phantom{-}0.018  & 10 & 13  & \Czhawi \\
& \phantom{-}0.014  & 10 & 14  & \Cforgei \\
& -0.012  & 11 & 14  & \Csheffi \\
& -0.012  & 11 & 14  & \Charv \\\hdashline[0.5pt/2pt]
\multirow{2}{*}{3}  & -0.078  & 15 & 16  & \Cthomsonii \\
& -0.083  & 15 & 16  & \Cforgeiii \\\hdashline[0.5pt/2pt]
\multirow{3}{*}{4}  & -0.152  & 17 & 19  & \Cadapt \\
& -0.185  & 17 & 19  & \Cthomsoni \\
& -0.186  & 17 & 19  & \Czhang \\\hdashline[0.5pt/2pt]
\multirow{2}{*}{5}  & -0.426  & 20 & 21  & \Cchen \\
& -0.457  & 20 & 21  & \Csheffii \\\hline
\end{tabular}
\quad
\begin{tabular}{ccr@{--}ll}
\multicolumn{5}{c}{\normalsize\bf Naturalness\Bstrut}\\\hline
\bf \#  & \bf TrueSkill  & \multicolumn{2}{c}{\bf Rank}  & \bf System \\\hline
1  & \phantom{-}0.211  & 1 & 1  & \Csheffii \\\hdashline[0.5pt/2pt]
\multirow{11}{*}{2}  & \phantom{-}0.171  & 2 & 3  & \Cslug \\
& \phantom{-}0.154  & 2 & 4  & \Cchen \\
& \phantom{-}0.126  & 3 & 6  & \Charv \\
& \phantom{-}0.105  & 4 & 8  & \Cnle \\
& \phantom{-}0.101  & 4 & 8  & \Ctgen \\
& \phantom{-}0.091  & 5 & 8  & \Cdangnt \\
& \phantom{-}0.077  & 5 & 10  & \Ctuda \\
& \phantom{-}0.060  & 7 & 11  & \Ctntnlgii \\
& \phantom{-}0.046  & 9 & 12  & \Cgong \\
& \phantom{-}0.027  & 9 & 12  & \Ctntnlgi \\
& \phantom{-}0.027  & 10 & 12  & \Czhang \\\hdashline[0.5pt/2pt]
\multirow{5}{*}{3}  & -0.053  & 13 & 16  & \Cthomsoni \\
& -0.073  & 13 & 17  & \Cslugalt~\emph{(late)} \\
& -0.077  & 13 & 17  & \Csheffi \\
& -0.083  & 13 & 17  & \Czhawii \\
& -0.104  & 15 & 17  & \Czhawi \\\hdashline[0.5pt/2pt]
\multirow{2}{*}{4}  & -0.144  & 18 & 19  & \Cforgei \\
& -0.164  & 18 & 19  & \Cadapt \\\hdashline[0.5pt/2pt]
\multirow{2}{*}{5}  & -0.243  & 20 & 21  & \Cthomsonii \\
& -0.255  & 20 & 21  & \Cforgeiii \\\hline
\end{tabular}
\end{center}

\caption{TrueSkill measurements of quality (left) and naturalness (right) for all primary systems (significance cluster number, TrueSkill value, range of ranks where the system falls in 95\% of cases or more, system name). Significance clusters are separated by a dotted line. 
System architectures are coded with colours and symbols: \textcolor{seqtoseq}{\symbseq seq2seq}, \textcolor{datadriven}{\symbdd other data-driven}, \textcolor{rules}{\symbrule rule-based}, \textcolor{templates}{\symbtempl template-based}.
}
\label{tab:human-results}
\end{table}

\begin{table}[tbp]
\begin{center}
\footnotesize
\setlength{\extrarowheight}{2pt}
\begin{tabular}{l>{\hspace{-0.2cm}}c>{\hspace{-0.1cm}}c>{\hspace{-0.1cm}}c>{\hspace{-0.1cm}}c}
\multicolumn{5}{c}{\normalsize\bf Human Ratings\Bstrut}\\\hline
\textbf{System} & \textbf{OK} & \bf A & \bf M & \bf A+M \\ \hline
\Cslug & 74\% & \phantom{0}8\% & 17\% & 1\% \\
\Cgong & 74\% & \phantom{0}6\% & 19\% & 1\% \\
\Cdangnt & 74\% & \phantom{0}9\% & 17\% & 0\% \\
\Ctuda & 74\% & 19\% & \phantom{0}7\% & 0\% \\
\Cthomsonii & 73\% & 10\% & 14\% & 3\% \\
\Csheffi & 72\% & \phantom{0}9\% & 18\% & 1\% \\
\Cslugalt & 70\% & 12\% & 18\% & 1\% \\
\Czhawii & 69\% & \phantom{0}8\% & 22\% & 1\% \\
\Ctgen & 69\% & \phantom{0}7\% & 23\% & 1\% \\
\Cforgei & 68\% & \phantom{0}9\% & 20\% & 3\% \\
\Ctntnlgi & 66\% & \phantom{0}7\% & 25\% & 1\% \\
\Ctntnlgii & 62\% & \phantom{0}9\% & 28\% & 1\% \\
\Czhawi & 61\% & \phantom{0}9\% & 28\% & 1\% \\
\Cforgeiii & 60\% & 10\% & 29\% & 1\% \\
\Cnle & 59\% & \phantom{0}8\% & 31\% & 2\% \\
\Charv & 53\% & \phantom{0}9\% & 35\% & 4\% \\
\Cthomsoni & 51\% & \phantom{0}8\% & 42\% & 0\% \\
\Cadapt & 51\% & 12\% & 33\% & 4\% \\
\Czhang & 43\% & \phantom{0}8\% & 49\% & 0\% \\
\Cchen & 27\% & 10\% & 62\% & 0\% \\
\Csheffii & 26\% & \phantom{0}9\% & 62\% & 3\% \\\hline
\end{tabular}
\hfill
\begin{tabular}{l>{\hspace{-0.2cm}}c>{\hspace{-0.1cm}}c>{\hspace{-0.1cm}}c>{\hspace{-0.1cm}}c>{\hspace{-0.1cm}}c}
\multicolumn{6}{c}{\normalsize\bf Automatic (pattern matching)\Bstrut}\\\hline
\textbf{System} & \textbf{OK} & \bf A & \bf M & \bf A+M & \bf SER\\ \hline
\Ctuda & \hspace{-1.5mm}100\% & 0\% & \phantom{0}0\% & \phantom{0}0\% & \phantom{0}0.00\% \\
\Csheffi & 93\% & 0\% & \phantom{0}5\% & \phantom{0}2\% & \phantom{0}1.08\% \\
\Cgong & 92\% & 4\% & \phantom{0}2\% & \phantom{0}2\% & \phantom{0}1.13\% \\
\Cforgei & 92\% & 0\% & \phantom{0}8\% & \phantom{0}0\% & \phantom{0}1.22\% \\
\Cslug & 91\% & 1\% & \phantom{0}4\% & \phantom{0}4\% & \phantom{0}1.26\% \\
\Cdangnt & 88\% & 0\% & 12\% & \phantom{0}0\% & \phantom{0}1.75\% \\
\Ctgen & 79\% & 3\% & 16\% & \phantom{0}2\% & \phantom{0}3.56\% \\
\Cslugalt & 78\% & 4\% & \phantom{0}9\% & \phantom{0}9\% & \phantom{0}3.56\% \\
\Czhawii & 76\% & 3\% & 20\% & \phantom{0}1\% & \phantom{0}3.68\% \\
\Ctntnlgi & 73\% & 1\% & 22\% & \phantom{0}4\% & \phantom{0}4.92\% \\
\Ctntnlgii & 71\% & 1\% & 28\% & \phantom{0}1\% & \phantom{0}6.04\% \\
\Czhawi & 70\% & 3\% & 25\% & \phantom{0}2\% & \phantom{0}5.12\% \\
\Cthomsonii & 66\% & 6\% & 23\% & \phantom{0}5\% & \phantom{0}5.45\% \\
\Cnle & 63\% & 3\% & 24\% & 10\% & \phantom{0}6.20\% \\
\Charv & 54\% & 2\% & 30\% & 14\% & 10.43\% \\
\Cadapt & 50\% & 3\% & 36\% & 10\% & 12.48\% \\
\Cthomsoni & 48\% & 0\% & 52\% & \phantom{0}0\% & 13.83\% \\
\Cforgeiii & 41\% & 0\% & 55\% & \phantom{0}3\% & 10.41\% \\
\Czhang & 27\% & 0\% & 73\% & \phantom{0}0\% & 14.80\% \\
\Cchen & 11\% & 0\% & 88\% & \phantom{0}1\% & 23.53\% \\
\Csheffii & \phantom{0}5\% & 0\% & 88\% & \phantom{0}6\% & 27.94\% \\\hline
\end{tabular}
\end{center}
\caption{Results of input MR coverage evaluation, with human ratings (left) and using an automatic pattern-matching script (right). Columns legend: \emph{OK} -- proportion of outputs covered perfectly, \emph{A} -- proportion of outputs with added information, \emph{M} -- with missed information, \emph{A+M} with both missed and added information, \emph{SER} -- slot/semantic error rate (see Section~\ref{sec:coverage}). The lists are sorted by the proportion of perfectly covered MRs. System architectures are colour-coded: \textcolor{seqtoseq}{seq2seq}, \textcolor{datadriven}{other data-driven}, \textcolor{rules}{rule-based}, \textcolor{templates}{template-based}.}
\label{tab:coverage}
\end{table}

\begin{table}[p]
\begin{center}
\scriptsize
\setlength{\extrarowheight}{2pt}
\begin{tabular}{ll>{\raggedright}p{8.8cm}r}\hline
\bf\# & \multicolumn{2}{l}{\bf MR / System name and output} & \bf Score \\
\hline
\multirow{9}{*}{1} & \multicolumn{2}{>{\raggedright}p{10cm}}{name[The Vaults], eatType[restaurant], food[French], priceRange[moderate], area[city centre], familyFriendly[yes], near[Raja Indian Cuisine]}\\\cdashline{2-4}[0.5pt/2pt]
 & \Cgong & The Vaults is a french restaurant in the city centre near Raja Indian Cuisine. It is kids friendly and has a moderate price range. & 100 \\ 
 & \Charv & The Vaults is a moderately priced French restaurant in the city centre near Raja Indian Cuisine. & 90 \\ 
 & \Cforgeiii & Family-friendly and moderate, The Vaults restaurant is located near Raja Indian Cuisine, in the city centre area. & 85 \\ 
 & \Czhang & The Vaults is a kid friendly French restaurant in the city centre near Raja Indian Cuisine . & 80 \\ 
 & \Csheffi & The Vaults is a moderately priced French restaurant near Raja Indian Cuisine in the city centre is. & 70\Bstrut \\\hline 
\multirow{6}{*}{2} & \multicolumn{2}{p{10cm}}{name[Cotto], eatType[coffee shop], near[The Bakers]} \\\cdashline{2-4}[0.5pt/2pt]
 & \Cthomsonii & Cotto is a coffee shop located near The Bakers. & 100 \\ 
 & \Cslugalt & Cotto is a coffee shop and is located near The Bakers & 97 \\ 
 & \Ctgen & Cotto is a coffee shop with a low price range. It is located near The Bakers. & 85 \\ 
 & \Cgong & Cotto is a place near The Bakers. & 85 \\ 
 & \Csheffii & Cotto is a pub near The Bakers. & 82\Bstrut \\\hline 
\multirow{6}{*}{3} & \multicolumn{2}{p{10cm}}{name[Clowns], eatType[coffee shop], customer rating[3 out of 5], near[All Bar One]} \\\cdashline{2-4}[0.5pt/2pt]
 & \Csheffi & Clowns is a coffee shop near All Bar One with a customer rating of 3 out of 5. & 100 \\ 
 & \Czhang & Clowns is a coffee shop near All Bar One with a customer rating of 3 out of 5 . & 100 \\ 
 & \Cforgeiii & Clowns is a coffee shop near All Bar One with a rating 3 out of 5. & 70 \\ 
 & \Czhawii & A coffee shop near All Bar One is Clowns. It has a customer rating of 3 out of 5. & 50 \\ 
 & \Csheffii & Clowns is a pub near All Bar One. & 20\Bstrut \\\hline 
\multirow{10}{*}{4} & \multicolumn{2}{p{10cm}}{name[The Cricketers], eatType[restaurant], customer rating[low], familyFriendly[no], near[Ranch]} \\\cdashline{2-4}[0.5pt/2pt]
 & \Cslug & The Cricketers is a restaurant near Ranch. It is not family friendly and has a low customer rating. & 72 \\ 
 & \Cslugalt & Located near Ranch, The Cricketers is a family-friendly restaurant that is not family-friendly and has a low customer rating. & 71 \\ 
 & \Cadapt & The Cricketers is a non - family - friendly restaurant located near the Ranch . It has a low customer satisfaction rating . & 68 \\ 
 & \Cforgei & The restaurant The Cricketers is near Ranch. The Cricketers, which does not welcome kids, has a low customer rating. & 65 \\ 
& \Ctuda & The Cricketers is a restaurant located near Ranch. It has a low customer rating. It is not family friendly. & 56 \\\hline\end{tabular} 
\end{center}

\vspace{-3mm}
\caption{Example system outputs with human rankings of quality and a detailed error analysis attempting to interpret the rankings.}\label{tab:quality-ratings}

\footnotesize\medskip
Each example is shown as ranked for quality by a single crowd worker. The raw RankME scores assigned by the crowd workers are shown; however, note that only relative ranks are used by the TrueSkill algorithm. The outputs within each example are sorted by the score for clarity.
For the purpose of error analysis, the rankings may be interpreted in the following way (note that quality rankings include both relevance and fluency):
\begin{enumerate}
\item \gong and \forgeiii verbalise all attributes but the latter is less fluent. \harv misses the family-friendliness, \zhang misses the price information. \sheffi misses family-friendliness and is not fluent.
\item \thomsonii and \slugalt provide perfect and fluent information but \slugalt misses the full stop. \gong does not specify the type of place while \tgen adds irrelevant price range information. \sheffii indicates a wrong venue type.
\item \sheffi and \zhang provide perfect and fluent information, \forgeiii is less fluent and \zhawii even less than that. \sheffii indicates a wrong venue type and misses the customer rating information.
\item \slug provides a perfect an fluent information. \slugalt is repetitive and \adapt was probably penalised for lack of detokenisation. \forgei and \tuda provide a complete information but are not very fluent.
\end{enumerate}
\end{table}

The results of human evaluation of quality and naturalness are provided in Table~\ref{tab:human-results}.
Using the RankME setup described in Section~\ref{sec:human-evaluation},
we collected 2,979 data points of partial system rankings for quality, where one data point corresponds to one MR and ranked
outputs of five randomly selected systems (see Table~\ref{tab:quality-ratings} for examples).
From these rankings, a set of 29,790 pairwise output comparisons were produced to be used by the TrueSkill algorithm.
This resulted in 1,418 pairwise comparisons per system.
For naturalness, 4,239 data points were collected, which resulted in 42,390 pairwise comparisons, and 2,018 comparisons per system.
For each of 630 MRs in the test set, 9.5 systems on average (with a maximum of 14) were compared based on both naturalness and quality of their outputs.
That is, using TrueSkill, we were able to reduce the number of required system comparisons to more than half.
The CrowdFlower task for collecting human evaluation data was running for 235~hours and cost USD~314 in total.

We produced the final ranking of all systems for both quality and naturalness using the TrueSkill algorithm with bootstrap resampling as described in Section~\ref{sec:human-evaluation}.
This resulted in clusters of systems with significantly different system rankings for both {\em naturalness} and {\em quality}.\footnote{Note that TrueSkill provides a relative ranking of a system in terms of its cluster and rank range (cf.~Section~\ref{sec:human-evaluation}), i.e.\ the numerical scores are not directly interpretable. Other systems in the same cluster are considered to show performance that is not significantly different. In other words: if a system is part of e.g.\ cluster 2, this system can be considered 2nd best, but it is sharing this position with all other systems in the cluster.}
In both cases, there are clear winning systems (i.e., the 1st cluster only has one member): \sheffii for {\em naturalness} and \slug for {\em quality}.
The 2nd clusters 
are quite large for both criteria 
-- they contain 13 and 11 systems, respectively, and they include the baseline \tgen system in both cases.

The results indicate that seq2seq systems dominate in terms of {\em naturalness} of their outputs, while most systems of other architectures score lower. The bottom cluster is filled with template-based systems. The winning \sheffii system is seq2seq-based, and the 2nd cluster mostly includes other seq2seq-based systems. The result also indicates that diversity-attempting systems are penalised in naturalness, i.e.\ \slugalt, \zhawi, \zhawii placed in the 3rd cluster; \adapt in the 4th.

The results for {\em quality}\footnote{Note that our definition of quality in Section~\ref{sec:human-evaluation} also includes semantic completeness and grammaticality.} are, however, more mixed in terms of architectures, with none of them clearly prevailing.
The 2nd, most populous cluster includes all different architecture types. The winner is the seq2seq-based system \slug. However, the bottom two clusters are also composed of seq2seq-based systems. This shows the importance of an explicit semantic control mechanism applied at decoding time in seq2seq systems: None of the systems in the bottom two clusters apply such mechanism, whereas all better ranking seq2seq systems do (cf.~Section~\ref{sec:semantic-control}).\footnote{While the \chen and \zhang systems do attempt to model the coverage of the input MR, they do not use explicit beam reranking based on MR coverage.} Note that this also includes the \sheffii system, which scored top for {\em naturalness}. With the exception of diversity-attempting \adapt, these systems tend to produce the shortest outputs (see Table~\ref{tab:systems-textual}), which indicates that they are penalised for not realising parts of the input MR too often (cf.~Section~\ref{sec:coverage}).

Finally, we computed the correlation of word-overlap metrics with the human judgements of both quality and naturalness for all the systems. All of the correlations are weak ($<0.2$, see Tables~\ref{tab:corr_qual} and~\ref{tab:corr_natur} in the Appendix), which confirms earlier findings of
\citet{Novikova:EMNLP2017} and explains the discrepancy between system performances in terms of automatic and human evaluation.

\subsection{Error Analysis: Input MR Coverage}\label{sec:coverage}

In order to clarify the mixed quality evaluation results, we attempted to estimate the number of semantic errors produced by the individual systems in two ways: First, we ran a specific crowdsourced evaluation of systems' coverage of the input MR, where crowd workers were asked to manually annotate missed and added information with respect to the input MR (see Table~\ref{tab:coverage}). We did not check for workers' correctness here, and thus we can expect some noise, but the annotations confirm that the systems rated low on quality, most of which also produce very short outputs, also correspond to the ones with the lowest proportion of perfectly covered MRs (\chen, \sheffii, \zhang, \thomsoni and \adapt).

Second, semantic errors were computed following \citet{reed_can_2018}, where we implemented a script to estimate the coverage automatically based on regular expression matching.\footnote{We based the patterns for the individual attribute-value pairs on \citet{reed_can_2018}'s script and manually enhanced them using the first 500 instances of the E2E development set.} This allowed us to produce an independent estimate of the proportion of outputs with missing or added information (see Table~\ref{tab:coverage}).
Following \citet{reed_can_2018}, we also computed the slot error rate (SER) using this pattern-matching approach and the following formula:\footnote{Note that the coverage and SER values produced by the script is only an estimate as the patterns for a given attribute-value pair will not cover all possible all correct ways to express it. This is different from \citet{wen:emnlp2015}'s computation of SER, where full delexicalisation allowed them to directly count placeholders in the output.}
\begin{equation}
\mbox{SER} = \frac{\mbox{\#\,missed} + \mbox{\#\,added} + \mbox{\#\,value errors} + \mbox{\#\,repetitions}}{\mbox{\#\,slots}}
\end{equation}
Here, \emph{missed} stands for slot values missing from the realisations, \emph{added} denotes additional information not present in the MR (hallucinations), \emph{value errors} denote correctly realised slots with incorrect values (e.g., specifying low price range instead of high), and \emph{repetitions} are values mentioned repeatedly in the outputs; \emph{slots} is the total number of slots/attributes in the test set. SER thus amounts to a proportion of erroneously realised slots.
While the absolute numbers for perfectly covered MRs are different from those estimated by humans, they mostly follow the same trend. The SER value is highly correlated with the proportion of perfectly covered MRs.

Both evaluations show that template- and rule-based systems, where MR coverage is implied by the architecture, mostly score high in this regard. However, \forgeiii, which uses template mining from training data, scores below average; here, some amount of noise was probably carried over from training data. \tuda, on the other hand, scores high in human ratings and even achieved perfect score by the automatic script (100\% perfect coverage), but this is partly given by its low diversity (cf.~Section~\ref{sec:results-text-metrics}) -- all its templates are probably covered well by the patterns. The results also show that some data-driven systems are able to achieve very good coverage (especially \sheffi, \gong and \slug, with SER estimates below 1.5\%), which confirms the efficacy of their respective semantic control approaches (see Section~\ref{sec:semantic-control}). Seq2seq systems without reranking (\chen, \sheffii, \zhang, \adapt, \thomsoni) score near the bottom of the list in both evaluations.

Both estimates also indicate that missing information is the most common type of problem, added (hallucinated) information occurs less frequently, but still poses a serious problem for utterance generation in task-based dialogue systems.\footnote{Note that this problem appears to be more general since it has also been reported in related fields, including image captioning \cite{rohrbach-etal-2018-object}, machine translation \cite{koehn_six_2017,lee2019hallucinations}, and question answering \cite{feng_pathologies_2018}.} It also appears that both problems are connected -- systems hallucinating less fre quently tend to miss information more often.

Finally, the scores show that attempts at diversity may hurt semantic accuracy. This is most apparent in \adapt, the most diverse system with no explicit semantic control mechanism. Other systems with diverse outputs, \forgeiii and \harv, also score lower on coverage.
In case of \forgeiii, this is due to the above-mentioned noise in the mined templates; \harv's reranking is probably less aggressive than others'.
On the other hand, \zhawi, \zhawii and especially \slugalt produce diverse outputs while maintaining good coverage  thanks to their very powerful semantic control mechanisms.

\subsection{Winning System}\label{sec:winner}

We consider the \slug system \citep{juraska_slug2slug:_2018}, a seq2seq-based ensemble system, 
as the overall winner of this challenge. It received high human ratings for both naturalness and quality, as well as for automatic word-overlap metrics. In contrast to vanilla seq2seq systems, \slug improves semantic coverage using a heuristic slot aligner
in combination with a data augmentation method producing partially aligned examples, which places it among the top-scoring systems in terms of MR coverage (cf.~Section~\ref{sec:coverage}).
\slug's only drawback is the relatively low output diversity; note that repetitive output is considered to be problematic for task-based dialogue systems.
A variant of the same system, \slugalt, provides much more output diversity at the cost of slightly lower quality ratings and MR coverage; it maintains higher quality and coverage scores than other diversity-attempting approaches.

While the \sheffii system \citep{chen_shefeld_2018}, a vanilla seq2seq setup, 
won in terms of naturalness, it
often does not realise all parts of the input MR, which severely affected its quality rating -- it placed in the last cluster, ranked 20th--21st out of 21. \sheffii's outputs also rank very low on complexity and diversity.

Furthermore, the \tgen baseline system turned out hard to beat. It ranked highest on average in word-overlap-based automatic metrics and placed in the 2nd cluster in both quality and naturalness (ranks 3--6 and 4--8 out of 21, respectively). \tgen also fared well (albeit not perfectly) in MR coverage evaluations. On the other hand, \tgen only scored in the middle of the pack on output diversity.

\subsection{Lessons Learnt and Future Directions}\label{sec:lessons}

We  attempt to formulate
some high-level ``lessons learnt" 
for developing future data-driven NLG systems based on the above results, while we acknowledge that our data is limited to a single domain, and  that comparisons are not strictly controlled, i.e.\ models vary in more than one aspect. 
\begin{itemize}
\item {\bf Semantic control}: For seq2seq-based systems, a strong semantic control of the generated content seems crucial -- beam reranking based on MR classification or heuristic alignments appears to work well while attention-only models perform poorly on our data. Correct semantics is regarded by users as more important than fluency \citep{belz:CL2009} and should be prioritised when training the models \cite[cf.\ also][]{reiter_does_2019}.

\item {\bf Open vocabulary}: For limited domains such as ours, delexicalisation of open-set attributes still seem to be the best approach. However, the systems of \harv and \nle show character-level models and copy mechanisms are viable alternatives. We believe that the low results of \chen, \zhang and \adapt are due to inferior semantic control, not open-vocabulary handling.

\item {\bf Complexity and diversity}: In general, hand-engineered systems seem to outperform neural systems in terms of output diversity and complexity (see Section~\ref{sec:results-text-metrics}); the most diverse outputs are produced by systems using templates mined from training data and data-driven systems with explicit diversity mechanisms.

Vanilla seq2seq-based systems produce the least diverse outputs: they are essentially probabilistic language models, which tend to settle for the most frequent phrasing, thus penalising length and favouring high-frequency word sequences.
Diversity in seq2seq models can be improved by data selection (\slugalt), diverse ensembling (\harv) or sampling from the generated beam \citep{wen:emnlp2015}.
In contrast, hand-engineered system authors can control the output complexity and diversity directly: here, \tuda's outputs are very repetitive as its set of handcrafted templates is small, while \forgeiii and \thomsonii with templates mined from data produce some of the most diverse outputs.

In general, any systems attempting output diversity need to impose strong semantic control mechanisms to maintain MR coverage.

\item {\bf Best method suggestion}: Rule-based methods work quite well for limited domains, such as ours. Low-effort handcrafting (as in \tuda) may lead to correct but repetitive outputs. Seq2seq models with semantic reranking emerge as the best data-driven option, in combination with controlling for diversity and using copy mechanisms to minimise preprocessing.
\end{itemize}

\section{Conclusion}\label{sec:conclusion}

This paper presents the findings of the first shared task on End-to-End Natural Language Generation for Spoken Dialogue Systems.
The aim of this challenge was to assess the capabilities of recent end-to-end, fully data-driven NLG systems, which can be trained from pairs of input meaning representations and corresponding texts, without the need for fine-grained semantic alignments.
In addition to attracting many participants, the challenge has substantially shaped current NLG research, as it has
influenced, inspired and motivated a number of recent studies outwith the original competition.

As part of this challenge, we have created a novel dataset for NLG benchmarking in the restaurant information domain, which is an order-of-magnitude bigger than any previous publicly available dataset for task-oriented NLG and has already been used and extended by multiple follow-up works since its original release.
We also provided one of the previous state-of-the art seq2seq-based NLG systems, \tgen \citep{Dusek:ACL16}, as a baseline for comparison.
The challenge received 62 system submissions by 17 different participating institutions. The systems submitted ranged from complex seq2seq-based setups with different additions to the architecture, over other data-driven methods and rule-based systems, to simple template-based ones. We evaluated all the entries in terms of five different automatic metrics. 20 primary submissions (as identified by the participants) were further evaluated using a novel, crowdsourced evaluation setup.
We also include a novel comparison of systems in terms of automatic textual metrics aimed to assess output complexity and diversity.
Our evaluation lets us include several general recommendations for future NLG system development.

In general, seq2seq-based systems produce very similar outputs (as measured by word-overlap, cf. Section \ref{sec:systems-similarity}), despite their different implementations. Seq2seq models tend to score high on word-overlap metrics and human evaluations of naturalness, while the scores for other data-driven, rule-based and template-based systems are lower. However, these other types of systems often score better in human evaluations of the overall quality.
While the winning \slug system is seq2seq-based, the results also demonstrated possible pitfalls of using seq2seq models:
\begin{enumerate}
\item Vanilla seq2seq models tend to produce short outputs of low diversity and syntactic complexity. Low diversity is especially problematic since it causes repetitive outputs in spoken dialogue systems.
\item Applying a strong semantic control mechanism during decoding is crucial to preserve the input meaning. The most common semantic mistake for systems is to miss out information. However, added information (hallucinations) is also closely linked. Both type of errors can have severe consequences for task-based dialogue systems, depending on the application domain.
\item Addressing these issues is challenging: Attempts to improve diversity can often result in lowered semantic accuracy and/or output naturalness.\footnote{This finding is in line with recent follow-up works to the challenge \cite{oraby_controlling_2018,oraby_curate_2019,balakrishnan_constrained_2019}, which suggests that explicit style supervision is needed to produce both diverse and accurate outputs.}
\end{enumerate}

In comparison, hand-engineered systems tend to produce more complex and diverse outputs and are able to reach high overall quality, but are mostly rated low on naturalness.
Note that similar findings have been reported by \citet{Wiseman:EMNLP17} for data-to-document generation.
This raises the general question regarding efficiency, costs, and performance of purely data-driven versus carefully hand-engineered NLG systems.

To facilitate further research in this domain, we have made the following data and tools freely available for download:
\begin{itemize}
\item The E2E NLG training dataset (including test set with human references),
\item A set of word-overlap-based metrics and scripts for running further textual metrics used for automatic evaluation in the challenge,
\item Outputs of the baseline \tgen system for the development set,
\item Outputs for the test set produced by the baseline and all participating systems,
\item the corresponding RankME ratings for quality and naturalness collected in the human evaluation campaign.
\end{itemize}
All can be accessed under the following URL:
\begin{center}
\url{http://www.macs.hw.ac.uk/InteractionLab/E2E/}
\end{center}

In future work, we aim to investigate additional evaluation methods for NLG systems, such as post-edits \citep{sripada2005evaluation}, or extrinsic evaluation, such as NLG's contribution to task success, e.g.\ \citep{Rieser:IEEE14,Gkatzia:acl16}.
We also intend to continue our work on automatic quality estimation for NLG \citep{dusek_referenceless_2017}, where the large amount of data obtained in this challenge allows a wider range of experiments than previously possible.

\section*{Acknowledgments}
This research received funding from the EPSRC projects  DILiGENt (EP/M005429/1) and  MaDrIgAL (EP/N017536/1) and Charles University project PRIMUS/19/SCI/10. The Titan Xp used for this research was donated by the NVIDIA Corporation.
The authors would like to thank Lena Reed and Shereen Oraby for help with computing the slot error rate.
We would also like to thank the anonymous reviewers for providing exceptionally helpful comments and Prof.~Ehud Reiter, whose blog\footnote{\url{https://ehudreiter.com/}} inspired some of this research.

\newpage

\bibliographystyle{elsarticle-harv}
\bibliography{references}

\onecolumn
\appendix

\section{Detailed Results}
\begin{table}[H]
\begin{center}
\vspace{-0.7cm}
\begin{adjustbox}{totalheight=\textheight-3\baselineskip}
\begin{tabular}{llccccccc}
\hline
\textbf{Submitter} & \textbf{System name} & \textbf{P?} & \bf BLEU & \bf NIST & \bf\scriptsize METEOR & \bf\scriptsize ROUGE-L & \bf CIDEr & \bf n.~avg.\\
\hline
Heriot-Watt Uni  & \tgen  & \checkmark  & 0.6593  & 8.6094  & 0.4483  & 0.6850  & 2.2338 & 0.5754 \\\hdashline[0.5pt/2pt]
B.\ Zhang, Xiamen Uni & \zhang  & \checkmark  & 0.6545  & 8.1840  & 0.4392  & 0.7083  & 2.1012 & 0.5661 \\\hdashline[0.5pt/2pt]
\multirow{7}{*}{\shortstack[l]{S.\ Chen, \\ Harbin Inst of Tech}} & \it abstract beam1  &   & 0.5854  & 5.4691  & 0.3977  & 0.6747  & 1.6391 & 0.4737 \\
  & \it abstract beam2  &   & 0.5916  & 5.9477  & 0.3974  & 0.6701  & 1.6513 & 0.4838 \\
  & \it abstract beam3  &   & 0.6150  & 6.8029  & 0.4068  & 0.6750  & 1.7870 & 0.5112 \\
  & \it abstract greedy  &   & 0.6635  & 8.3977  & 0.4312  & 0.6909  & 2.0788 & 0.5666 \\
  & \it non-abstract beam2  &   & 0.5860  & 6.1602  & 0.3833  & 0.6619  & 1.6133 & 0.4817 \\
  & \it non-abstract beam3  &   & 0.6088  & 6.9790  & 0.3899  & 0.6628  & 1.7015 & 0.5059 \\
  & \chen & \checkmark  & 0.5859  & 5.4383  & 0.3836  & 0.6714  & 1.5790 & 0.4685 \\\hdashline[0.5pt/2pt]
\multirow{3}{*}{\shortstack[l]{Zurich Uni of \\ Applied Sciences}}   & \it base  &   & 0.6544  & 8.3391  & 0.4448  & 0.6783  & 2.1438 & 0.5652 \\
  & \zhawi  & \checkmark  & 0.5864  & 8.0212  & 0.4322  & 0.5998  & 1.8173 & 0.5205 \\
  & \zhawii  & \checkmark  & 0.6004  & 8.1394  & 0.4388  & 0.6119  & 1.9188 & 0.5314 \\\hdashline[0.5pt/2pt]
\multirow{3}{*}{Pompeu Fabra Uni} & \forgei  & \checkmark  & 0.4207  & 6.5139  & 0.3685  & 0.5437  & 1.3106 & 0.4231 \\
  & \it 2  &   & 0.4113  & 6.3293  & 0.3686  & 0.5593  & 1.2467 & 0.4194 \\
  & \forgeiii  & \checkmark  & 0.4599  & 7.1092  & 0.3858  & 0.5611  & 1.5586 & 0.4547 \\\hdashline[0.5pt/2pt]
\multirow{6}{*}{Sheffield NLP}  & \sheffi & \checkmark  & 0.6015  & 8.3075  & 0.4405  & 0.6778  & 2.1775 & 0.5537 \\
  & \it primary1 var2  &   & 0.6233  & 8.1751  & 0.4378  & 0.6887  & 2.2840 & 0.5591 \\
  & \it primary1 var3  &   & 0.5690  & 8.0382  & 0.4202  & 0.6348  & 2.0956 & 0.5275 \\
  & \it primary1 var4  &   & 0.5799  & 7.9163  & 0.4310  & 0.6670  & 2.0691 & 0.5353 \\
  & \sheffii  & \checkmark  & 0.5436  & 5.7462  & 0.3561  & 0.6152  & 1.4130 & 0.4462 \\
  & \it primary2 var2  &   & 0.5356  & 7.8373  & 0.3831  & 0.5513  & 1.5825 & 0.4824 \\\hdashline[0.5pt/2pt]
\multirow{4}{*}{\shortstack[l]{HarvardNLP \\\& Adapt}}  & \it support 1  &   & 0.6581  & 8.5719  & 0.4409  & 0.6893  & 2.1065 & 0.5712 \\
  & \it support 2  &   & 0.6618  & 8.6025  & 0.4571  & 0.7038  & 2.3371 & 0.5833 \\
  & \it support 3  &   & 0.6737  & 8.6061  & 0.4523  & 0.7084  & 2.3056 & 0.5851 \\
  & \harv  & \checkmark  & 0.6496  & 8.5268  & 0.4386  & 0.6872  & 2.0850 & 0.5673 \\\hdashline[0.5pt/2pt]
\multirow{4}{*}{\shortstack[l]{H.\ Gong, \\ Harbin Inst of Tech}} & \gong  & \checkmark  & 0.6422  & 8.3453  & 0.4469  & 0.6645  & 2.2721 & 0.5631 \\
  & \it 1  &   & 0.6396  & 8.3111  & 0.4466  & 0.6620  & 2.2272 & 0.5604 \\
  & \it 3  &   & 0.6395  & 8.3127  & 0.4457  & 0.6628  & 2.2442 & 0.5607 \\
  & \it 4  &   & 0.6395  & 8.3127  & 0.4457  & 0.6628  & 2.2442 & 0.5607 \\\hdashline[0.5pt/2pt]
\multirow{3}{*}{Adapt Centre}   & \adapt  & \checkmark  & 0.5092  & 7.1954  & 0.4025  & 0.5872  & 1.5039 & 0.4738 \\
   & \it temperature 0.9  &   & 0.5573  & 7.7013  & 0.4154  & 0.6130  & 1.8110 & 0.5074 \\
   & \it temperature 1.0  &   & 0.5265  & 7.3991  & 0.4095  & 0.5992  & 1.6488 & 0.4880 \\\hdashline[0.5pt/2pt]
\multirow{2}{*}{$<$anonymous$>$}   & \it combined  &   & 0.2921  & 4.7690  & 0.2515  & 0.4361  & 0.6674 & 0.3047 \\
   & \it primary  &  & 0.4723  & 6.1938  & 0.3170  & 0.5616  & 1.2127 & 0.4183 \\\hdashline[0.5pt/2pt]
\multirow{3}{*}{Naver Labs Europe}  & \nle & \checkmark  & 0.6534  & 8.5300  & 0.4435  & 0.6829  & 2.1539 & 0.5696 \\
  & \it second  &   & 0.6669  & 8.5388  & 0.4484  & 0.6991  & 2.2239 & 0.5781 \\
  & \it third  &   & 0.6676  & 8.5416  & 0.4485  & 0.6991  & 2.2276 & 0.5784 \\\hdashline[0.5pt/2pt]
\multirow{2}{*}{\shortstack[l]{UC Santa Cruz \\-- Slug2Slug}}   & \slug  & \checkmark  & 0.6619  & 8.6130  & 0.4454  & 0.6772  & 2.2615 & 0.5744 \\
  & \slugalt \it (late)  & \checkmark  & 0.6035  & 8.3954  & 0.4369  & 0.5991  & 2.1019 & 0.5378 \\\hdashline[0.5pt/2pt]
\multirow{10}{*}{\shortstack[l]{Thomson Reuters\\ NLG}}  & \it 1 model 11 post  &   & 0.6536  & 8.3293  & 0.4550  & 0.6805  & 2.1050 & 0.5665 \\
  & \it 2 model 13 post  &   & 0.6562  & 8.3942  & 0.4571  & 0.6876  & 2.1706 & 0.5715 \\
  & \it 3 beam 5 model 11 post\hspace{-5mm}  &   & 0.6805  & 8.7777  & 0.4462  & 0.6928  & 2.3195 & 0.5858 \\
  & \it 4 beam 5 model 13 post\hspace{-5mm}  &   & 0.6742  & 8.6590  & 0.4499  & 0.6983  & 2.3018 & 0.5837 \\
  & \it 5 submission 6  &   & 0.6208  & 8.0632  & 0.4417  & 0.6692  & 2.1127 & 0.5499 \\
  & \it 6 submission 4 beam  &   & 0.6201  & 8.0938  & 0.4419  & 0.6740  & 2.1251 & 0.5516 \\
  & \it 7 submission 4  &   & 0.6182  & 8.0616  & 0.4417  & 0.6729  & 2.0783 & 0.5494 \\
  & \it 8 train only  &   & 0.4111  & 6.7541  & 0.3970  & 0.5435  & 1.4096 & 0.4336 \\
  & \thomsoni & \checkmark  & 0.6336  & 8.1848  & 0.4322  & 0.6828  & 2.1425 & 0.5563 \\
  & \thomsonii  & \checkmark  & 0.4202  & 6.7686  & 0.3968  & 0.5481  & 1.4389 & 0.4372 \\\hdashline[0.5pt/2pt]
\multirow{6}{*}{\shortstack[l]{UC Santa Cruz \\-- TNT NLG}}  & \tntnlgi  & \checkmark  & 0.6561  & 8.5105  & 0.4517  & 0.6839  & 2.2183 & 0.5729 \\
  & \it sys1 model1  &   & 0.6476  & 8.4301  & 0.4508  & 0.6795  & 2.1233 & 0.5666 \\
  & \tntnlgii  & \checkmark  & 0.6502  & 8.5211  & 0.4396  & 0.6853  & 2.1670 & 0.5688 \\
  & \it sys2 model1  &   & 0.6606  & 8.6223  & 0.4439  & 0.6772  & 2.1997 & 0.5728 \\
  & \it sys2 model2  &   & 0.6563  & 8.5482  & 0.4482  & 0.6835  & 2.1953 & 0.5725 \\
 & \it sys2 model3  &   & 0.3681  & 6.6004  & 0.3846  & 0.5259  & 1.5205 & 0.4181 \\\hdashline[0.5pt/2pt]
VNU-HCM Uni of IT  & \dangnt  & \checkmark  & 0.5990  & 7.9277  & 0.4346  & 0.6634  & 2.0783 & 0.5395 \\\hdashline[0.5pt/2pt]
Tech Uni Darmstadt  & \tuda  & \checkmark  & 0.5657  & 7.4544  & 0.4529  & 0.6614  & 1.8206 & 0.5215 \\\hline
\end{tabular}
\end{adjustbox}
\end{center}
\vspace{-0.3cm}
\caption{Full list of E2E challenge submissions with automatic metric scores (primary systems are indicated in the ``P?'' column; the column ``n.~avg.'' shows an average of all metrics normalised into the 0-1 range, cf.~Table~\ref{tab:primary-systems-wbms}).}
\vspace{-1.5cm}
\label{tab:all-systems-wbms}
\end{table}

\begin{figure}[p]
\begin{center}
\includegraphics[width=0.49\linewidth]{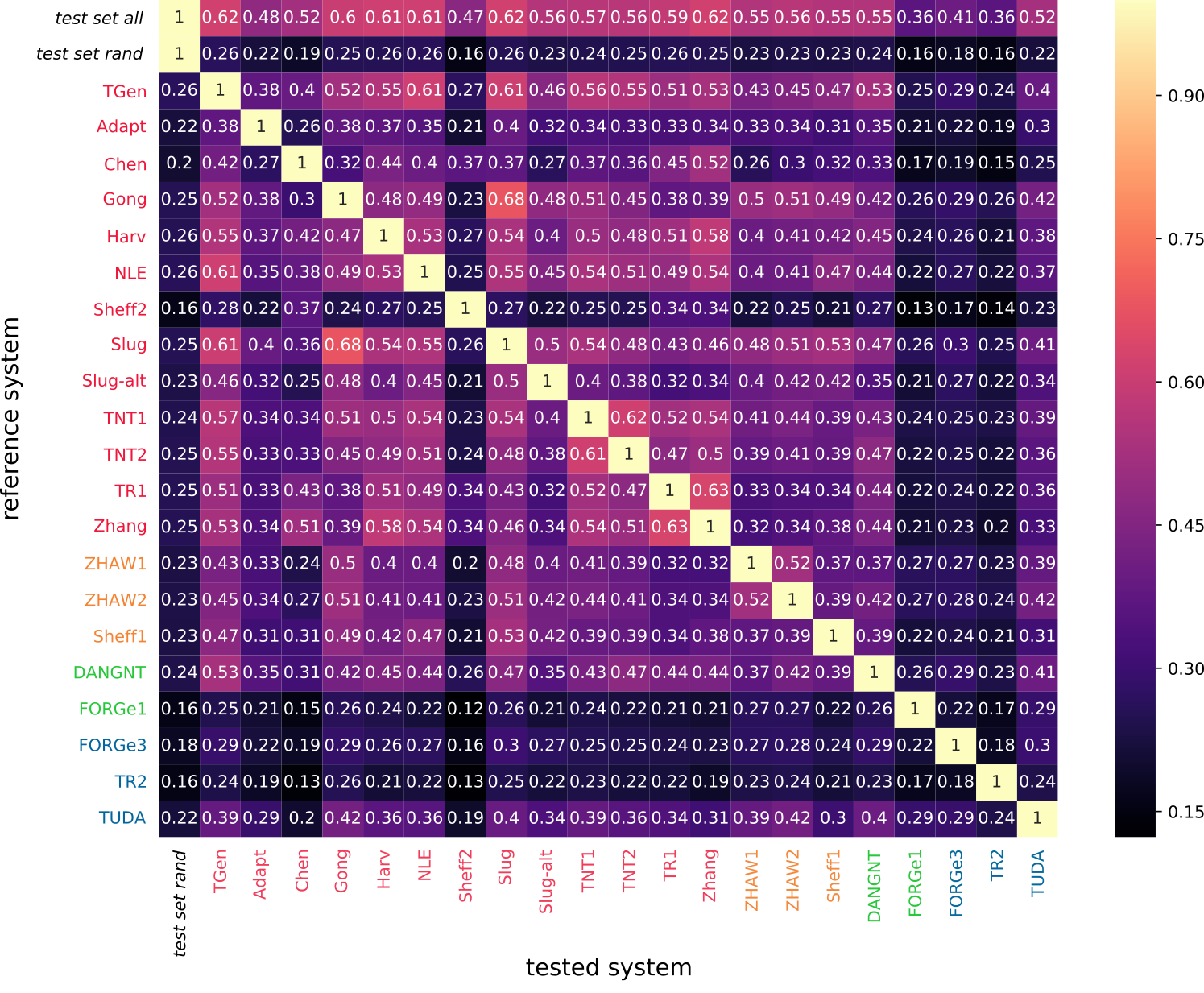}\hfill
\includegraphics[width=0.49\linewidth]{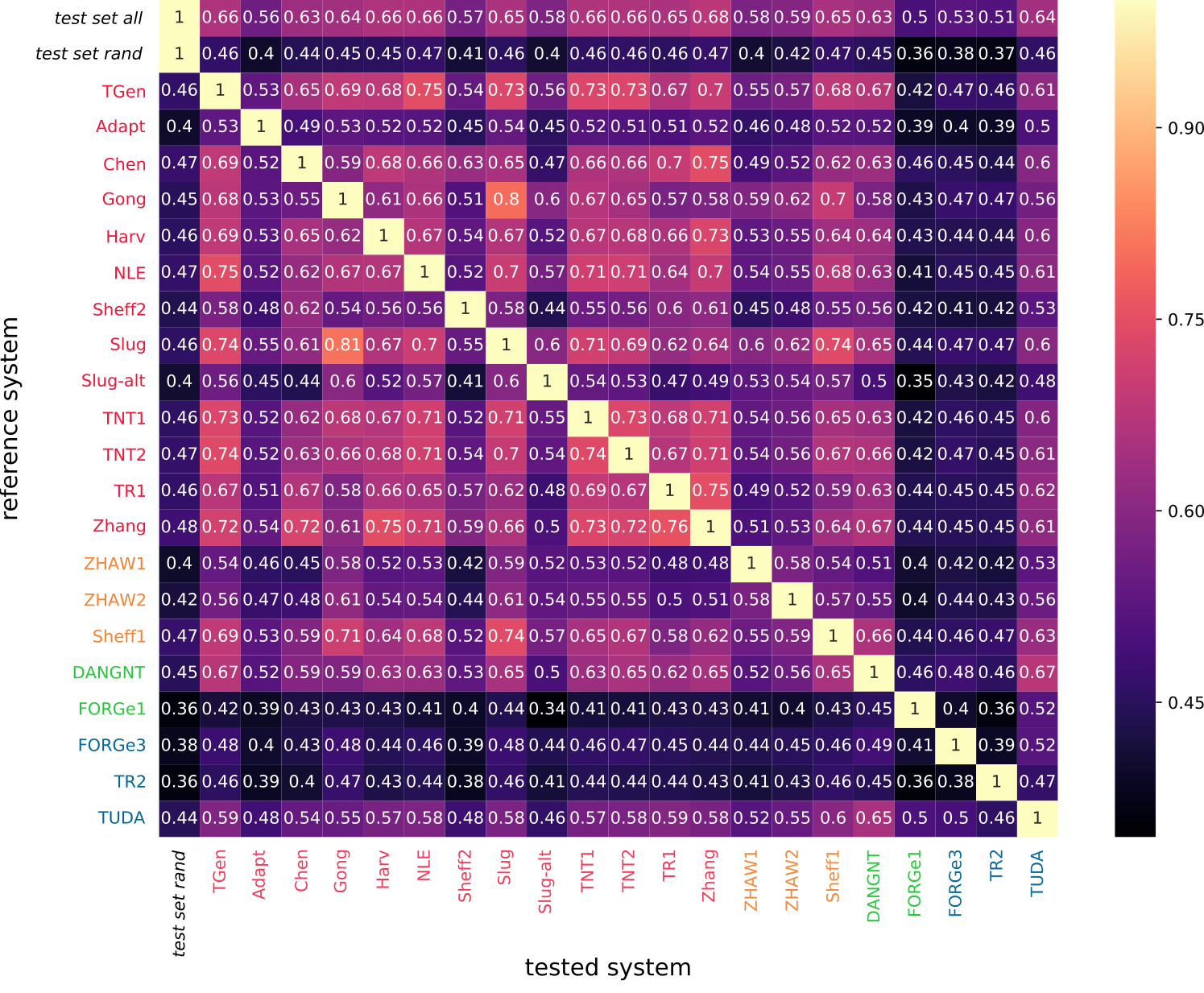} \\
\vspace{5mm}
\includegraphics[width=0.49\linewidth]{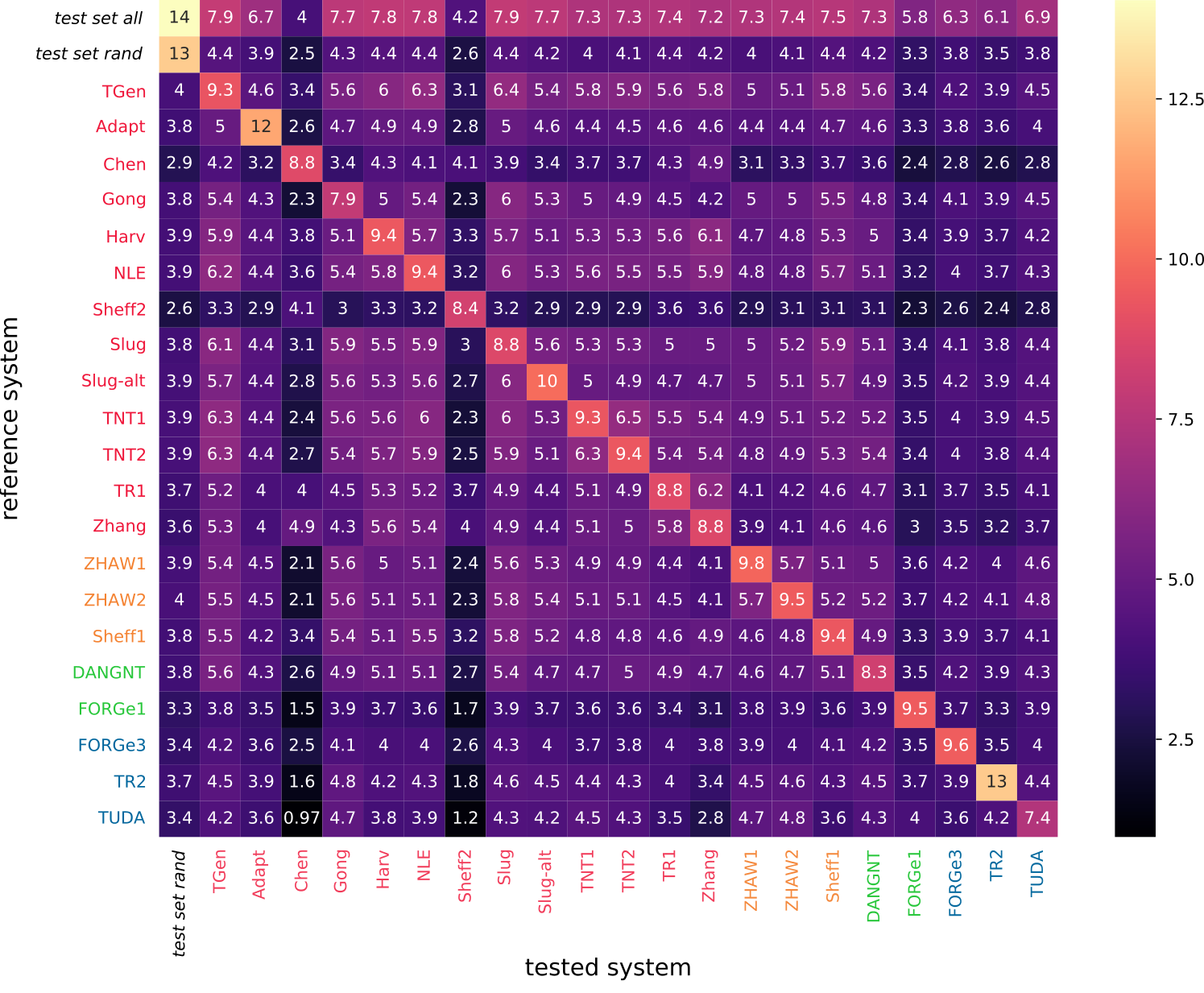}\hfill
\includegraphics[width=0.49\linewidth]{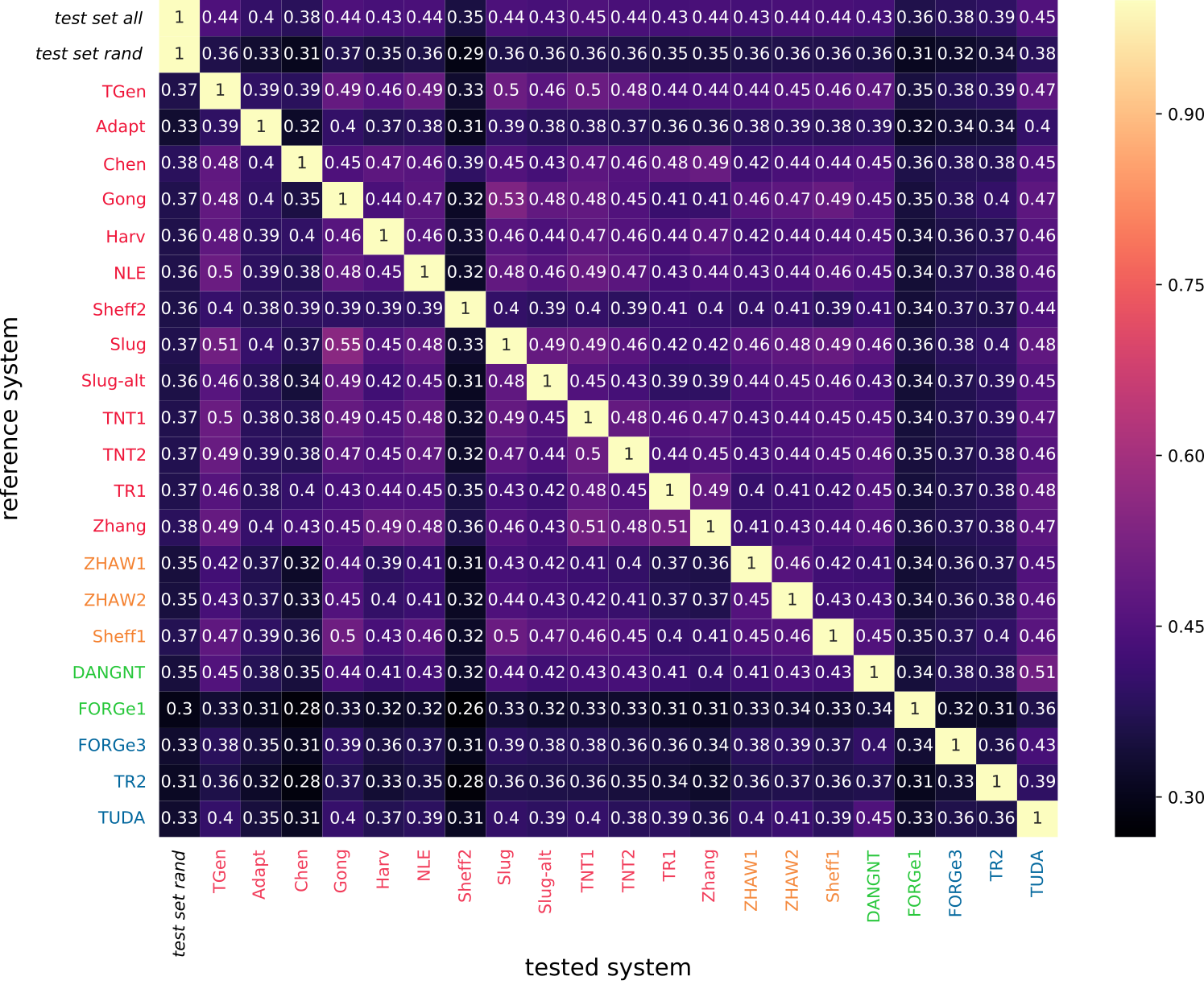} \\
\vspace{5mm}
\includegraphics[width=0.49\linewidth]{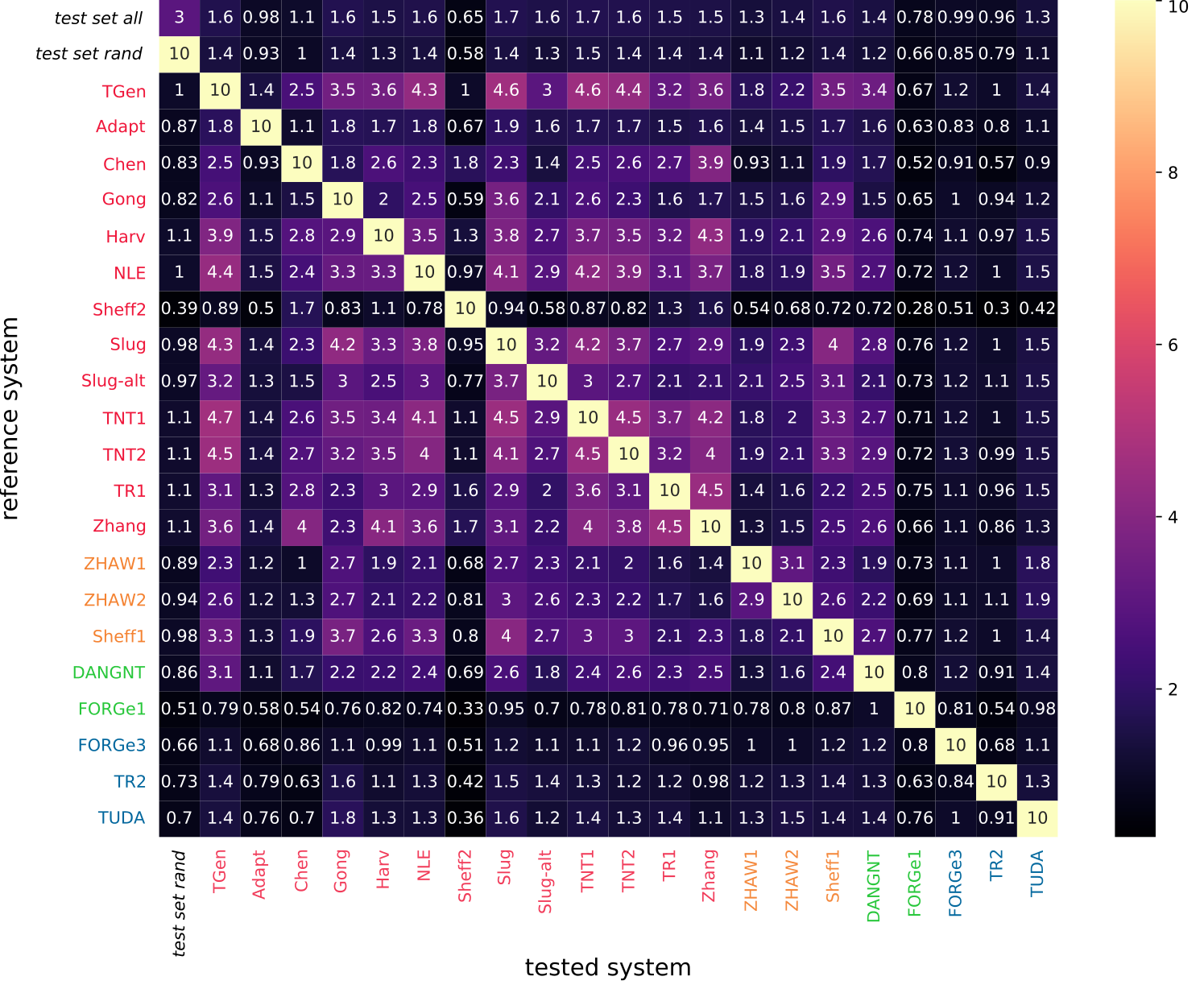}
\end{center}
\caption{Similarity of the systems' outputs as measured by automatic metrics (left-to-right, top-to-bottom: \bleu, \nist, \meteor, \rouge and \cider), where one of the systems is used as a reference. Systems within the graphs are sorted by their architecture. For comparison, we also include metrics values against the full test set with multiple huamn references and against a single (randomly chosen) test set human reference.}\label{fig:metrics-heatmap-all}
\end{figure}

\begin{figure}[tbp]
\centering
\includegraphics[width=0.9\linewidth]{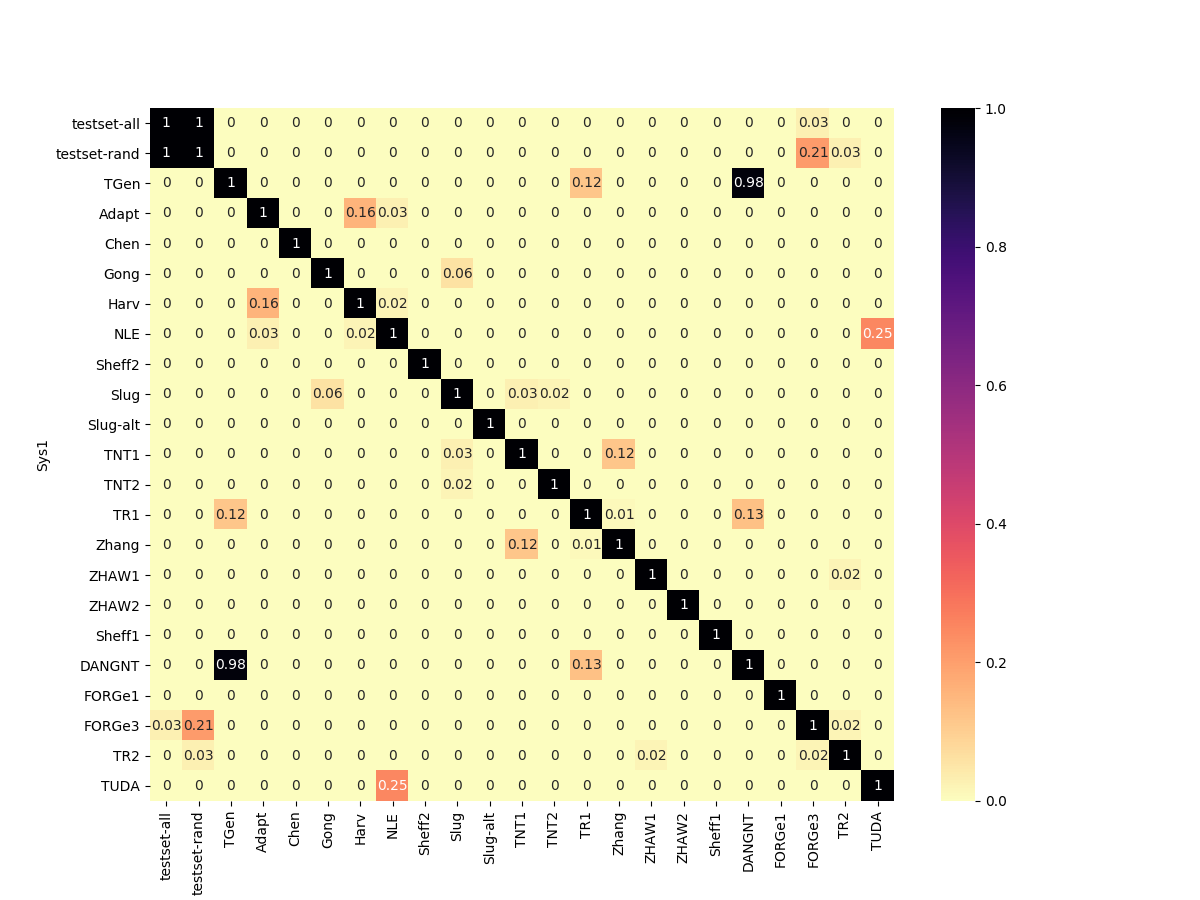}
\caption{$p$-values of the Kolmogorov-Smirnov test for discrete distributions \citep{arnold_nonparametric_2011}, evaluating significance of differences between systems in terms of syntactic complexity of their output. Bright color indicates statistically significant difference ($p<0.05$).}
\label{tab:diff_dlevels}
\end{figure}

\begin{figure}[tbp]
\centering
\includegraphics[width=0.9\linewidth]{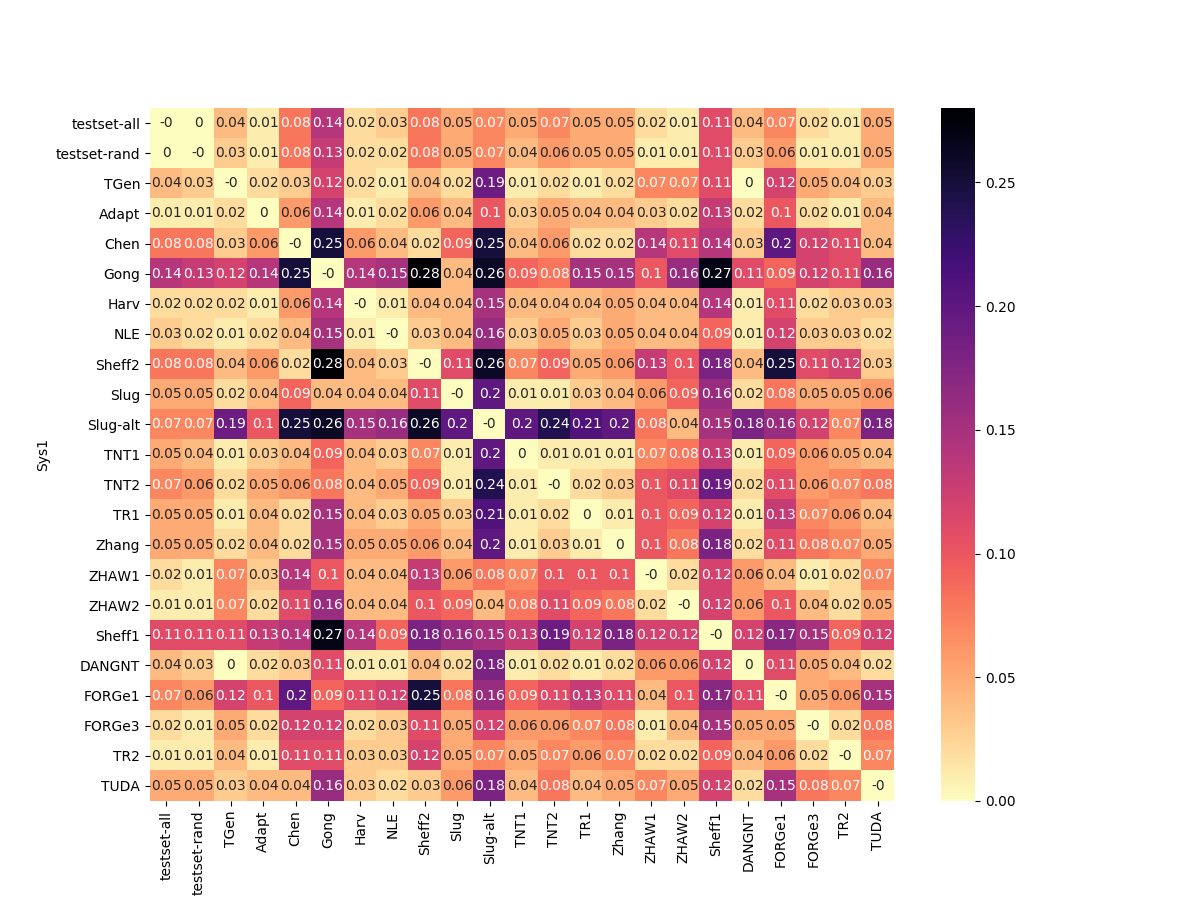}
\caption{Similarities between systems, calculated using Bhattacharyya distance. Darker color indicates greater distance, i.e. more different systems.}
\label{tab:dist_dlevels}
\end{figure}

\begin{table}[tpb]
\centering
\setlength{\extrarowheight}{1pt}
\footnotesize
\begin{tabular}{lFFFFF}\hline
\textbf{System name} & \mcC{\textbf{\bleu}} & \mcC{\textbf{\nist}} & \mcC{\textbf{\meteor}} & \mcC{\textbf{\rouge}} & \mcC{\textbf{\cider}} \\ \hline
\Ctgen & 0.08 & 0.00 & 0.03 & 0.04 & 0.03 \\
\Cadapt & 0.05 & 0.05 & 0.05 & 0.09 & 0.09 \\
\Cchen & 0.03 & -0.03 & -0.06 & 0.01 & -0.08 \\
\Cdangnt & 0.06 & -0.11 & 0.02 & 0.06 & 0.08 \\
\Cforgei & -0.06 & -0.13* & -0.03 & 0.06 & 0.05 \\
\Cforgeiii & 0.03 & 0.02 & -0.03 & -0.02 & 0.04 \\
\Cgong & 0.08 & 0.10 & 0.00 & 0.08 & -0.02 \\
\Charv & 0.04 & 0.05 & 0.02 & 0.06 & -0.09 \\
\Cnle & 0.07 & 0.08 & 0.10 & 0.05 & 0.11 \\
\Csheffi & 0.07 & 0.11 & 0.01 & 0.03 & -0.12* \\
\Csheffii & -0.11 & -0.08 & -0.08 & -0.04 & 0.02 \\
\Cslug & 0.02 & 0.08 & -0.07 & 0.03 & -0.05 \\
\Cslugalt & -0.02 & -0.03 & 0.05 & 0.02 & 0.01 \\
\Cthomsoni & 0.15* & 0.13* & 0.15* & 0.15* & 0.02 \\
\Cthomsonii & 0.02 & 0.00 & 0.08 & 0.07 & 0.05 \\
\Ctntnlgi & -0.07 & -0.01 & -0.08 & -0.02 & -0.08 \\
\Ctntnlgii & 0.04 & 0.07 & 0.02 & 0.03 & -0.02 \\
\Ctuda & -0.02 & -0.03 & 0.13 & -0.04 & -0.01\\
\Czhang & 0.03 & 0.01 & 0.03 & 0.00 & -0.04 \\
\Czhawi & 0.05 & 0.00 & 0.05 & 0.08 & 0.01 \\
\Czhawii & \bf 0\bf .\bf 16* & 0.12* & 0.09 & 0.10 & 0.02 \\\hline
\end{tabular}
\caption{Pearson correlation between automatic metrics and human scores of naturalness. ``*'' denotes statistical significance at $p<0.05$ level, bold denotes the highest value.}
\label{tab:corr_natur}
\end{table}

\begin{table}[tbp]
\centering
\footnotesize
\setlength{\extrarowheight}{1pt}
\begin{tabular}{lFFFFF}\hline
\textbf{System name} & \mcC{\textbf{\bleu}} & \mcC{\textbf{\nist}} & \mcC{\textbf{\meteor}} & \mcC{\textbf{\rouge}} & \mcC{\textbf{\cider}} \\ \hline
\Ctgen & -0.08 & -0.10 & -0.05 & -0.02 & -0.05 \\
\Cadapt & 0.11* & 0.09 & 0.07 & 0.10 & 0.10 \\
\Cchen & 0.07 & \bf 0\bf .\bf 19* & 0.00 & 0.04 & 0.08 \\
\Cdangnt & -0.06 & 0.00 & -0.08 & -0.07 & -0.13* \\
\Cforgei & -0.01 & -0.02 & 0.06 & -0.01 & 0.08 \\
\Cforgeiii & 0.00 & -0.01 & 0.07 & 0.14* & 0.04 \\
\Cgong & 0.01 & -0.03 & -0.03 & 0.02 & -0.01 \\
\Charv & 0.01 & 0.15* & 0.05 & -0.01 & 0.16* \\
\Cnle & 0.09 & 0.03 & 0.05 & 0.09 & 0.15* \\
\Csheffi & 0.08 & 0.02 & 0.06 & 0.11 & 0.05 \\
\Csheffii & 0.07 & 0.17* & 0.10 & 0.00 & 0.16* \\
\Cslug & 0.04 & 0.00 & 0.03 & -0.01 & 0.06 \\
\Cslugalt & 0.07 & 0.01 & 0.01 & 0.02 & 0.08 \\
\Cthomsoni & 0.07 & 0.16* & -0.04 & 0.02 & 0.08 \\
\Cthomsonii & 0.04 & -0.02 & 0.09 & 0.05 & 0.08 \\
\Ctntnlgi & 0.04 & -0.04 & -0.01 & 0.00 & 0.02 \\
\Ctntnlgii & 0.05 & 0.06 & 0.06 & 0.05 & 0.08 \\
\Ctuda & 0.01 & 0.01 & 0.02 & 0.05 & 0.01 \\
\Czhang & 0.02 & 0.16* & 0.05 & 0.01 & 0.10 \\
\Czhawi & -0.11 & -0.05 & 0.00 & -0.10 & -0.10 \\
\Czhawii & 0.05 & 0.02 & 0.01 & 0.09 & 0.05 \\\hline
\end{tabular}
\caption{Pearson correlation between automatic metrics and human scores of quality. ``*'' denotes statistical significance at $p<0.05$ level, bold denotes the highest value.}
\label{tab:corr_qual}
\end{table}

\end{document}